\documentclass[10pt,twocolumn,letterpaper]{article}

\usepackage[pagenumbers]{cvpr}   

\usepackage{color, colortbl}
\usepackage{multirow}
\definecolor{cvprblue}{rgb}{0.21,0.49,0.74}
\usepackage[pagebackref,breaklinks,colorlinks,allcolors=cvprblue]{hyperref}

\usepackage{xr-hyper}
\usepackage{algorithm}
\usepackage{algpseudocode}
\usepackage{amsmath,bm}
\usepackage{bbm}

\newcommand{\qq}{\mathbf{q}}

\newcommand{\pp}{\mathbf{p}}                     
\newcommand{\xx}{\mathbf{x}}                     
\newcommand{\yy}{\mathbf{y}^{\text{\tiny{ohe}}}} 
\newcommand{\rr}{\mathbf{r}}                     
\newcommand{\cc}{\mathbf{c}}                     
\newcommand{\kmarg}{\mathbf{m}}                
\newcommand{\nmarg}{\mathbf{u_{(N+M)}}}                
\newcommand{\uu}{\mathbf{u}}                     
\newcommand{\real}{\mathbb{R}}                   
\newcommand{\vv}{\mathbf{v}}                     
\newcommand{\ttt}{\mathbf{t}}                    
\newcommand{\temp}{\tau}                         
\newcommand{\vl}{\bm{l}}                         
\newcommand{\vq}{\bm{q}}                         
\newcommand{\domain}[1]{$\mathcal{D}_\text{#1}$} 
\newcommand{\Q}{\mathbf{Q}}                      
\newcommand{\Ent}{\mathcal{H}}                   

\definecolor{darkblue}{rgb}{0.21,0.49,0.74}
\newcommand{\appensecref}[1]{\textcolor{darkblue}{Appendix~\ref{#1}}}

\definecolor{myblue}{RGB}{38, 145, 199}  
\definecolor{mygreen}{RGB}{38, 199, 149}

\newcommand{\Hquad}{\hspace{0.2em}} 
\newcommand{\mypar}[1]{\noindent\textbf{#1}\Hquad}

\usepackage[euler]{textgreek}
\usepackage{arydshln}
\usepackage{marvosym}

\usepackage{graphicx}
\usepackage{color} 

\definecolor{Gray}{gray}{0.9}
\definecolor{Better}{rgb}{0.18, 0.407, 0.266}
\definecolor{Worse}{rgb}{0.35, 0.35, 0.35}

\definecolor{ours_color}{gray}{0.9}

\newcommand{\our}{\cellcolor{ours_color}}
\newcommand{\best}[1]{\textbf{#1}}
\newcommand{\imp}[1]{$_{{\textbf{\textcolor{Better}{#1}}}}$}
\newcommand{\wor}[1]{$_{{\textbf{\textcolor{Worse}{#1}}}}$}
\newcommand{\nc}[1]{ \textcolor{Worse}{\textbf{#1}}}

\usepackage[accsupp]{axessibility}

\def\paperID{16001} 
\def\confName{CVPR}
\def\confYear{2025}

\title{Conformal Prediction for Zero-Shot Models}

\author{
Julio Silva-Rodríguez\textsuperscript{\Letter} \qquad
Ismail {Ben Ayed} \qquad
Jose Dolz \\
ÉTS Montréal\\
\Letter {\tt \small julio-jose.silva-rodriguez@etsmtl.ca}
}

\begin{document}

\maketitle
\thispagestyle{empty}


\begin{abstract}

Vision-language models pre-trained at large scale have shown unprecedented adaptability and generalization to downstream tasks. Although its discriminative potential has been widely explored, its reliability and uncertainty are still overlooked. In this work, we investigate the capabilities of CLIP models under the split conformal prediction paradigm, which provides theoretical guarantees to black-box models based on a small, labeled calibration set. In contrast to the main body of literature on conformal predictors in vision classifiers, foundation models exhibit a particular characteristic: they are pre-trained on a one-time basis on an inaccessible source domain, different from the transferred task. This domain drift negatively affects the efficiency of the conformal sets and poses additional challenges. To alleviate this issue, we propose Conf-OT, a transfer learning setting that operates transductive over the combined calibration and query sets. Solving an optimal transport problem, the proposed method bridges the domain gap between pre-training and adaptation without requiring additional data splits but still maintaining coverage guarantees. We comprehensively explore this conformal prediction strategy on a broad span of 15 datasets and three non-conformity scores. Conf-OT provides consistent relative improvements of up to 20$\%$ on set efficiency while being $\times$15 faster than popular transductive approaches. We make the code available \footnote{\url{https://github.com/jusiro/CLIP-Conformal}}.

\end{abstract}



\section{Introduction}
\label{main:section_introduction}

Deep learning is currently undergoing a paradigm shift with the emergence of large-scale vision-language models (VLMs), such as CLIP \cite{radford2021learning}. These models, which are trained on a massive amount of paired language and image data leveraging contrastive learning techniques, have demonstrated unprecedented zero-shot capabilities on a wide array of downstream visual tasks, including classification \cite{radford2021learning,novack2023chils}, object detection \cite{linlearning,minderer2024scaling}, segmentation \cite{luo2023segclip,liang2023open} or image synthesis \cite{rombach2022high}, among many others. Inspired by the transferability power of VLMs, many efforts have focused on improving the discriminative performance of CLIP during adaptation to downstream tasks \cite{zhou2022coop,gao2021clip,zhang2021tip,kgcoop23,yu2023task,lin2023crossmodal,lp24,clap24}. 

Following their remarkable performance on general computer vision tasks, VLMs, and more particularly CLIP, are becoming increasingly popular in safety-critical scenarios, such as autonomous driving and medical imaging \cite{liang2022effective,liu2023clip,shakeri2024few,FLAIR}. Therefore, ensuring the reliability of model predictions is paramount for the safe deployment of these models in real-world applications, particularly considering their increasing adoption. Nevertheless, this crucial aspect has often been overlooked in the literature, with only a handful of recent works exploring the uncertainty of CLIP predictions from a calibration standpoint \cite{tpt,sals,tuempirical,oh2023towards}. 

Albeit popular, confidence calibration methods lack theoretical guarantees of the actual model performance. For example, these cannot estimate the most likely output (or set of outputs) and provide a verified probability of such prediction being correct. A principled solution is to quantify the uncertainty via conformal prediction (CP) frameworks \cite{learning_ny_transduction,transductive_ci,trans_conf_machine,inductive_ci,vovk_book}, which has experienced a growing interest in more traditional machine learning models. CP provides confidence guarantees by yielding prediction sets that contain the correct label with a desired coverage level, e.g., they can ensure that the true category will be part of the predictive sets, on average, 95\% of the time. Particularly, \textit{split conformal prediction} \cite{inductive_ci,vovk_book} provides a practical scenario to incorporate such marginal guarantees to \textit{black-box} models by leveraging a small \textit{calibration set}, which is assumed to be, at least, exchangeable with respect to test data \cite{vovk_book}. With the growing interest in the trustworthiness of machine learning systems, many works have explored CP on classical image classification benchmarks \cite{raps,conftr,ding2024class,saps}, including ImageNet \cite{deng2009imagenet} or CIFAR \cite{krizhevsky2009learning} datasets. For example, these works have focused on proposing novel criteria to create the predictive sets (non-conformity scores in conformal prediction literature) with improved efficiency \cite{lac}, adaptiveness \cite{aps,raps}, or conditional coverage \cite{ding2024class}.

\begin{figure*}[!ht]
\setlength{\tabcolsep}{1.0pt}
    \begin{center}
        \begin{tabular}{cccc}
         \multicolumn{2}{c}{\includegraphics[width=.50\linewidth]{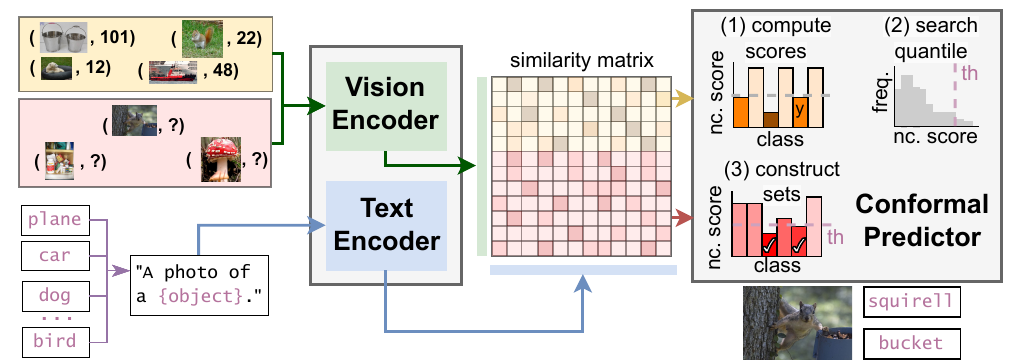}} & \hspace{2mm}
         \includegraphics[width=.23\linewidth]{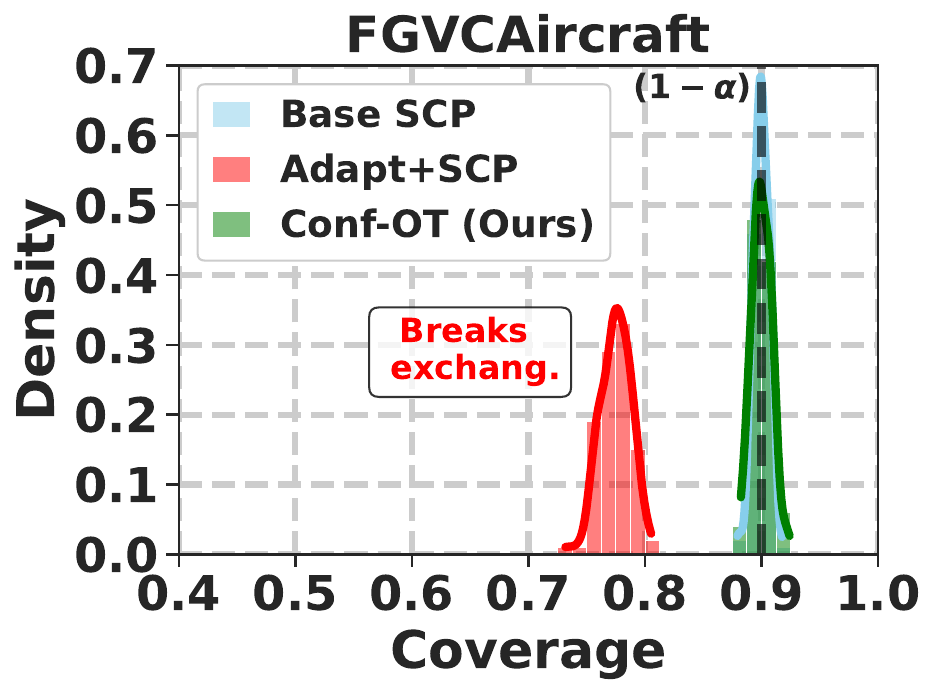} & 
          \includegraphics[width=.23\linewidth]{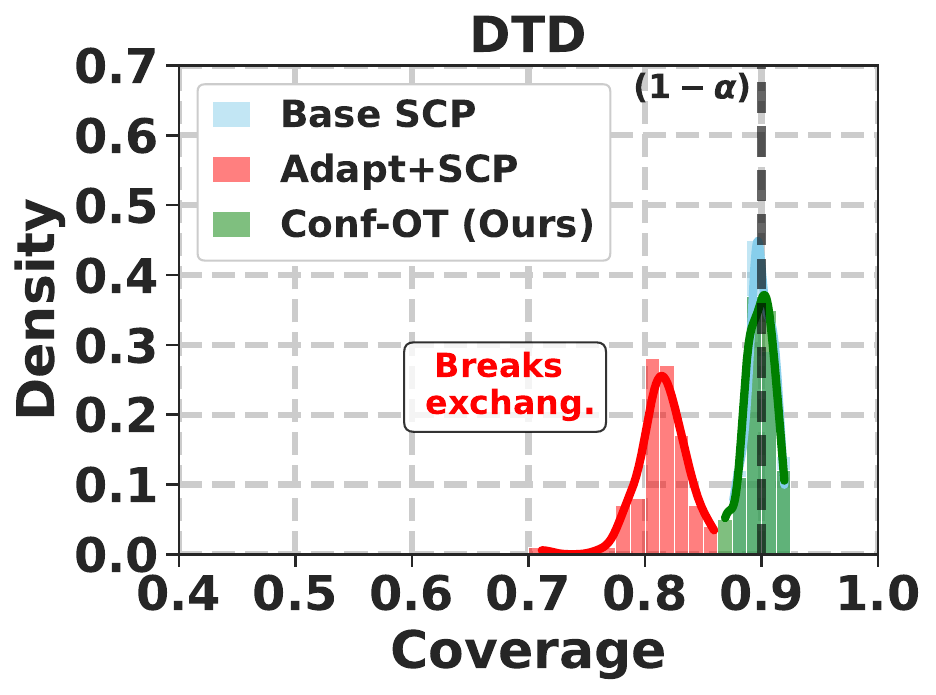} \\

        \multicolumn{2}{c}{(a) Conformal prediction in VLMs} & 
        \multicolumn{2}{c}{(b) Transfer learning and exchangeability.}
\\

        \end{tabular}
        \vspace{-2mm}
        \caption{\textbf{How to transfer black-box VLMs without breaking exchangeability?} In this work, we explore split conformal prediction (SCP) for VLMs (\textit{see} (a)) to provide trustworthiness guarantees. These zero-shot models typically undergo adaptation to enhance their performance. However, leveraging the SCP calibration data for adaptation breaks the exchangeability assumption \cite{vovk_book}, which produces miss-coverage during inference (\textit{see} (b)). We propose a transductive, unsupervised transfer to overcome such challenges, coined Conf-OT.}
        \label{fig:inductive}
    \end{center}
    \vspace{-5mm}
\end{figure*}

Building on these observations, this work explores how conformal prediction can be integrated into vision-language models to enhance their reliability while maintaining a competitive performance. Indeed, vision-language foundation models are promising \textit{black-box} predictors, as evidenced by the existing literature \cite{radford2021learning,Menon2023}. Nevertheless, their zero-shot predictive performance depends on the source data distribution and concept frequency \cite{udandarao2024zeroshot}. Thus, they usually require an adaptation stage when severe domain gaps exist in the target tasks w.r.t. the pre-training data assembly. This adaptation can be performed with efficient linear probing solutions \cite{lin2023crossmodal,clap24}. However, this situation is problematic in the conformal prediction framework. In particular, if adjusting these classifiers using a few labeled examples, e.g., in a calibration set gathered for conformal prediction, the testing data scores may not be exchangeable w.r.t. calibration. Hence, the theoretical guarantees of conformal prediction will not hold, as illustrated in \cref{fig:inductive}(b). This motivates the following question: \textit{can the performance of VLMs in conformal settings be improved via transfer learning without additional data sources beyond the calibration set?\footnote{An additional labeled few-shot adaptation set could be introduced, thus keeping calibration data exchangeable to future queries. Nevertheless, demanding more labeled data might be \textit{unrealistic} in practice, e.g., in critical scenarios such as detecting rare, low-prevalence diseases \cite{fundusrare,skinrare,shakeri2024few}.}} \\[-1.0ex]

\noindent The main contributions of this paper can be summarized as:

\begin{itemize}
    \item We introduce the split conformal prediction framework for large-scale pre-trained vision-language models, providing trustworthiness guarantees on the zero-shot predictions based on a small labeled calibration set.

    \item In contrast to the main corpus of recent literature in computer vision, which explores the CP framework using specialized models, VLMs are pre-trained on a generalist, inaccessible source domain, different from the downstream task and data distribution. To address this challenge, we propose Conf-OT, an unsupervised transfer learning framework that reduces the domain gap while maintaining coverage guarantees. Concretely, the proposed transductive strategy aims to solve the optimal transport problem on the joint calibration and query text-driven similarity matrix, producing a code assignment that respects the marginal properties of the target distribution. 

    \item We provide extensive experiments to assess the performance of popular non-conformity scores atop black-box predictions produced by CLIP models, including 15 popular image classification benchmarks. The results demonstrate the effectiveness of Conf-OT to improve the set size efficiency and class-conditional coverage.

    \item Notably, upon the standard black-box conformal prediction paradigm, Conf-OT substantially outperforms recent transductive methods in the literature — even in the discriminative aspect — yet being a \textit{training-free} approach, which requires minimal computational overhead.

\end{itemize}


\section{Related work}
\label{main:section_rw}

\mypar{Zero-shot and transfer learning in VLMs.} Contrastive VLMs such as CLIP models exhibit outstanding generalization capabilities \cite{radford2021learning}, and enable zero-shot image classification without adaptation \cite{Menon2023}, despite being notably more accurate when concepts are represented during pre-training \cite{udandarao2024zeroshot}. The latter limitation has directed the community toward developing data-efficient adaptation techniques, usually under the few-shot paradigm \cite{zhou2022coop,gao2021clip,zhang2021tip,kgcoop23,lin2023crossmodal,lp24,clap24}. Particularly, efficient black-box Adapters, which only require embedded representations \cite{gao2021clip,zhang2021tip,lin2023crossmodal,lp24,clap24}, are playing a pivotal role in this topic. The best results are obtained through advanced linear probing techniques that combine text-driven class prototypes with few-shot visual information \citep{lin2023crossmodal,lp24,clap24}. As stated earlier, the reliability of these models remains less explored, with just a few recent works assessing the calibration aspect of VLMs \cite{tpt,sals}. In contrast, our work focuses on a more principled framework for uncertainty quantification of VLMs outputs. To the best of our knowledge, there has been limited exploration of predictive uncertainty in vision-language models from a conformal prediction standpoint.

\mypar{Transductive adaptation for image classification.} A direction to improve pre-trained models consists of leveraging the shared information of unlabeled test data, so-called \textit{transduction} — in contrast to its more extended \textit{inductive} counterpart, which makes independent predictions for each new data point. The first setting usually reports notable performance gains over the second \cite{tim}, at the cost of additional test-time computation. Several transductive methods fine-tune the whole encoder \cite{wang2021tent,Dhillon2020A}, whereas others promote lightweight black-box adaptation \cite{tim,ease,trans_dirichlet,zanella2024boosting}. The latter usually adjusts the class-wise prototypes in the feature space by exploiting mutual information on the query set \cite{tim,tim_followup}, or optimal transport \cite{ease}. Regarding VLMs, its transductive adaptation has been less explored, with only a few recent works \cite{trans_dirichlet,zanella2024boosting} focusing on the discriminative aspect. For example, \cite{trans_dirichlet} develops a solution for small tasks modeling the target data through a Dirichlet probability distribution, while TransCLIP \cite{zanella2024boosting} integrates a KL-divergence into a GMM clustering that encourages the predictions not to deviate from the textual prototypes. 

\mypar{Conformal prediction in vision classifiers.} Conformal prediction is a framework for uncertainty quantification that produces statistically valid predictive regions \cite{learning_ny_transduction,transductive_ci,trans_conf_machine,inductive_ci,pmlr-v25-vovk12}. This work focuses on \textit{split conformal prediction} \cite{inductive_ci,vovk_book}, a resource-efficient, practical setting that allows conformalizing any black-box classifier. In particular, given a trained model that outputs logit predictions, it assumes access to a fresh labeled calibration set exchangeable \cite{vovk_book} with testing data. This data is exploited to find a confidence-specific threshold from a non-conformity score, which is later employed for creating predictive sets with theoretical guarantees over such confidence level. To this end, different scores have been proposed \cite{lac,aps,raps}. Least Ambiguous Classifier, a.k.a. LAC \cite{lac} creates predictive sets by directly using the raw class probabilities. Adaptive Prediction Sets (APS) \cite{aps} computes the score by accumulating the sorted softmax values in descending order, and its regularized extension RAPS \cite{raps} tames the tail by enforcing small sets integrating explicit penalties.

Prior art on split conformal prediction has been validated on vision classification tasks \cite{aps,raps,conftr,confts,conftr_24,einbinder2022training}. Nevertheless, these evaluations assume narrow scenarios, using specialized (only-vision) models, usually trained with a large corpus of data in-distribution w.r.t. calibration/test. This focus significantly differs from the current emerging paradigm in vision, driven by large-scale pre-training using VLMs, which are transferred to a broad corpus of downstream tasks \cite{radford2021learning,zhou2022coop,gao2021clip}. Note that this zero-shot setting does not affect the coverage guarantees of split conformal prediction, which are distribution-free, but might hamper the efficiency and, hence, the usability of the produced sets.

\mypar{Transduction in conformal prediction} has been classically linked to \textit{full conformal prediction} \cite{learning_ny_transduction,transductive_ci,trans_conf_machine,vovk_trans_conf}. However, this framework — see \appensecref{rel_work_extended} — differs from the split setting addressed in our work. Particularly, it does not consider access to a calibration set. Instead, it evaluates each test data-label pair conformity by resorting to multiple model fits. It is worth noting that leveraging test data distribution is not exclusive to full conformal methods. For example, \cite{2024transconf} explores a transfer learning scenario with a domain shift between train and calibration/test under the split conformal prediction umbrella. Particularly, the authors study a transductive strategy to reduce the domain gap during training, from which we draw inspiration. Nevertheless, the scenario in \cite{2024transconf} assumes access to the source training data and requires training the base model, which drastically differs from our focus on foundation models.

\section{Background}
\label{preliminaries}

\subsection{Zero-shot models}
\label{background_zeroshot}

\mypar{Contrastive vision-language pre-training.} Large-scale VLMs, such as CLIP \cite{radford2021learning}, are trained on large heterogeneous datasets to encode similar representations between paired image and text information. CLIP comprises a vision encoder, $f_\theta(\cdot)$, and a text encoder, $f_\phi(\cdot)$. These encoders project data points into an $\ell_{2}$-normalized $D$-dimensional shared embedding space, yielding the corresponding visual, ${\vv} \in \real^{D \times 1}$, and text, ${\ttt} \in \real^{D \times 1}$, embeddings.

\mypar{Zero-shot inference.} For a particular image classification task, CLIP-based models can provide predictions based on the similarity between category prompts, i.e., text descriptions for the new categories, and testing images. Given a set of $K$ classes and an ensemble of $J$ text prompts for each one, $\{\{\ttt_{kj}\}_{j=1}^{J}\}_{k=1}^{K}$, a common practice is to obtain a zero-shot prototype for each target category by computing the center of the $\ell_2$-normalized text embeddings for each class, $\ttt_{k}=\frac{1}{J}\sum_{j=1}^{J}\ttt_{kj}$. Thus, for a given query image, the zero-shot prediction, $\hat{\pp}=(\hat{p}_k)_{1 \leq k \leq K}$, is obtained from the softmax cosine similarity between its vision embedding, $\vv$, and category prototypes $\ttt_{k}$: 
\begin{align}
\label{eq:zs_prediction}
    \phantom{,}\hat{p}_k
    = \frac
    {\exp( \vv^\top \ttt_{k} / \temp^{\text{\tiny{CLIP}}})}
    {\sum_{i=1}^{K} \exp( \vv_i^\top \ttt_i / \temp^{\text{\tiny{CLIP}}})},
\end{align}

\noindent where $l_k= (\vv^\top \ttt_k / \temp^{\text{\tiny{CLIP}}})$ are the logits. Note that $\vv^\top\ttt$ is the dot product, equivalent to cosine similarity, as vectors are $\ell_2$-normalized. Thus, logits are similarity measures for each sample to the textual class prototypes scaled with $\temp^{\text{\tiny{CLIP}}}$, a temperature parameter learned during the pre-training.

\subsection{Conformal prediction}
\label{background_conformal}

\mypar{Preliminaries.} Let us denote the black-box scores for an input image space of a zero-shot model, e.g., CLIP outputs in \cref{eq:zs_prediction}, as $\mathcal{X}\subset \real^{1 \times K}$. Also, we denote their corresponding label space, $\mathcal{Y}=\{1, 2, ..., K\}$, and $(\xx,y)$ as a random data pair sampled from a joint distribution $\mathcal{P}_{\mathcal{XY}}$.

\mypar{Split conformal inference.} To provide trustworthiness on the outputs of a machine learning model, conformal prediction \cite{vovk_book} aims to produce predictive sets containing the ground truth label with a user-specified probability. Formally, the goal is to construct a set-valued mapping $\mathcal{C}:\mathcal{X}\rightarrow 2^{K}$, from a model output such that:
\begin{align}
\label{eq:marginal}
    \mathcal{P}(Y\in C(\xx)) \geq 1-\alpha,
\end{align}

\noindent where $\alpha \in (0, 1)$ denotes the desired error rate (e.g., $10\%$), and $C(\xx) \subset \mathcal{Y}$ is the prediction set. This is denoted as the \textit{coverage guarantee}, and is \textit{marginal} over $\mathcal{XY}$. 

\textit{Split conformal prediction} \cite{inductive_ci} assumes a practical setting for black-box models, enabling deploying coverage guarantees for any predictor \cite{Lei2018}. First, it grants access to a labeled calibration subset \domain{cal}$=\{(\xx_i,y_i)\}_{i=1}^{N}$. Second, it assumes that the test data, \domain{test}$=\{(\xx_i)\}_{i=N+1}^{N+M}$, are \iid or exchangeable \cite{vovk_book} fresh data points, not used for training.
\textit{First}, the split conformal prediction process starts by defining a non-conformity score $s_{i}=\mathcal{S}(\xx_i,y_i)$ for each calibration sample, where $s_{i}$ is a measure of deviation between an example and the training data, which we will specify later. \textit{Second}, the 1-$\alpha$ quantile of the non-conformity score is determined from calibration data, which will serve as a confidence threshold to satisfy a given coverage:
\small
\begin{align}
\label{eq:threshold}
    \hat{s} = \text{inf}
    \biggl[
    s : 
    \frac{ |i\in \{1,...,N\}: s_{i} \leq s| } { N }
    \geq 
    \frac{ \lceil (N+1)(1-\alpha) \rceil } { N }  
    \biggr].
\end{align}
\normalsize
\textit{Third}, for each testing sample, the non-conformity score for each label is calculated. The prediction set comprises labels whose non-conformity score falls within $\hat{s}$:
\begin{align}
\label{eq:inference}
    \mathcal{C}(\xx) = \{ y \in \mathcal{Y} : \mathcal{S}(\xx,y) \leq \hat{s} \}.
\end{align}

\mypar{Non-conformity scores.} Different criteria have been proposed, aiming to produce small (\aka \textit{efficient}) sets but able to model \textit{adaptiveness}, e.g., larger predictive for uncertain test points. For the first, LAC \cite{lac} tends to produce the smallest possible predictive sets, while for the latter, adaptive scores such as APS \cite{aps} and RAPS \cite{raps} are popular options in vision. These are introduced in \appensecref{appendix_scores}. 

\mypar{Black-box setting.} The standard split conformal prediction setting deployed in vision tasks \cite{raps,confts,ding2024class} usually takes as input the raw logits produced by the base model, $\vl_i$, and contemplates the possibility of controlling the sharpness of its distribution, e.g., by using temperature scaling \cite{raps,confts} before producing softmax scores, i.e., $\pp_{i}=\sigma_{k}(\vl_i/\temp)$, being $\sigma(\cdot)_{k}$ the softmax activation, and $\temp$ a temperature parameter. Once the classwise probabilities are obtained, these are used as input for computing non-conformity scores, i.e., $\xx_i=\pp_{i}$.
 
\section{Proposed solution}
\label{main:section_approach}

\subsection{Conformal prediction in zero-shot models}

\mypar{Motivation.} Prior art in conformal prediction for vision \cite{raps,conftr,ding2024class,saps} assumes access to specialized black-box models pre-trained on a training subset drawn from the same data distribution as calibration and test. Nevertheless, this scenario is \textit{unrealistic} when transferring cutting-edge foundation models, particularly zero-shot VLMs. These are pre-trained on multiple tasks from inaccessible source data that differs from the target domain.

\mypar{Transfer learning setting.} Let us assume a scenario in which a black-box model from a source domain, \domain{train}, produces logits for a set of target categories. Also, for a new task, there exists a labeled calibration set \domain{cal}$=\{(\vl_i,y_i)\}_{i=1}^{N}$, and unlabeled testing data, \domain{test}$=\{(\vl_i)\}_{i=N+1}^{N+M}$, and we aim to create conformal predictive sets. Importantly, \domain{cal} and \domain{test} are exchangeable distributions from a target domain, which are different from \domain{train}.

\mypar{Problem statement.} The first measure to produce efficient sets is to learn a transfer function from the source to the target domains. One naive option would involve leveraging \domain{cal} supervision to adapt the black-box outputs, e.g., following few-shot adaptation literature \cite{gao2021clip,clap24}. Nonetheless, it is crucial to consider the final conformal predictive scenario. As shown in \cref{fig:inductive}(b), modeling the logit distribution to maximize the likelihood of label assignment using such labels would break the exchangeability assumptions.

\subsection{Transductive conformal prediction}

We propose a transfer learning strategy, which is: \textit{i}) \textit{unsupervised}, i.e., does not directly rely on label supervision, and \textit{ii}) \textit{transductive}, i.e., calibration and test (queries in the transductive literature) data points are jointly transferred. Thus, the proposed setting avoids introducing any distributional shifts that could potentially break the exchangeability assumption required in conformal prediction.

\mypar{Optimal transport for transfer learning.} We leverage well-established knowledge in optimal transport (OT) \cite{ot_book,cuturi2013sinkhorn} to learn a joint mapping from source to target domain in the label assignments in an unsupervised manner. Such technical choice is motivated by the capabilities of OT to produce assignments that respect, for instance, a given label-marginal distribution — estimated from the calibration set — thus reducing potential domain drifts. Our approach, coined \textbf{Conf-OT}, is detailed in \cref{alg:confot}, and we describe each component below.

\vspace{1mm}
\noindent \textit{Learning objective.} Let us consider the combined calibration and test sets, which are integrated into a similarity matrix, $\mathbf{S}\in \real^{K\times(N+M)}$. Each column represents the similarities to each category prototype, i.e., the logits of a given sample, $\vl_i$, extracted as detailed in \cref{background_zeroshot}. Our goal is to find the joint probabilities matrix, $\Q \in \real^{K \times (N+M)}_+$, typically referred to as \textit{codes}, which maximizes the similarity assignment. Note that each column in $\Q$, i.e., $\qq_{i}$, is the prototype assignment for each sample. To achieve this, we propose to cast this task as an \textit{optimal transport problem}, introducing marginal constraints for the expected target distribution. Formally, the search problem can be defined as:
\begin{equation}
\label{eq:ot_1}
    \max_{\Q \in \mathcal{Q}} \; tr(\Q^\top \mathbf{S}),
\end{equation}
where the matrix $\Q$ is relaxed to be an element of the \textit{transportation polytope}:
\begin{equation}\label{polytope}
\mathcal{Q}=\{ \Q \mid \Q \mathbf{1}_{(N+M)} = \kmarg, \Q^\top \mathbf{1}_K = \nmarg \}, 
\end{equation}
such that $\mathbf{1}_{(\cdot)}$ denotes a column vector of ones, and $\mathbf{u}_{(\cdot)}$ an uniform distribution, being $(\cdot)$ the input vector length. In this element, $\kmarg$ and $\nmarg$ determine the marginal distributions expected in the target domain. First, $\nmarg = \frac{1}{(N+M)} \mathbf{1}_{(N+M)}$, is the sample-conditional marginal distribution, which is expected to be uniform to distribute the total similarity across all data points evenly. On the other hand, $\kmarg$, is the label-marginal distribution of the class assignments. Despite using a uniform distribution $\kmarg=\uu_K=\frac{1}{K} \mathbf{1}_K$ has provided satisfactory results on different computer vision tasks \cite{caron2018deep,asanoself}, in our scenario, we constrain the solution to respect the observed label-marginal distribution on the calibration set, such that $\kmarg=\frac{1}{N}\sum_1^N \yy_i$, where $\yy_i$ is the one-hot encoding of $y_i$.

\vspace{1mm}
\noindent \textit{Optimization.} The objective in \cref{eq:ot_1} is a linear program. However, its optimization is not straightforward, particularly regarding the computational complexity, exacerbated by increasing data points and categories. To alleviate this issue and provide a fast adaptation strategy, we resort to the Sinkhorn algorithm \cite{cuturi2013sinkhorn}, which integrates an entropic constraint, enforcing a simple structure on the optimal regularized transport. Hence, the optimization problem becomes:
\begin{equation}
\label{eq:main}
\max_{\Q \in \mathcal{Q}} \; tr(\Q^\top \mathbf{S}) + \varepsilon \Ent(\Q),
\end{equation}
where \( \Ent(\Q) \) is the entropy, $\Ent(\Q) = - \sum_{ki} q_{ki} \log q_{ki}$, such that $q$ are elements of $\Q$, and $\varepsilon$ controls its weight. Now, the soft codes $\Q^*$ are the solution of the problem presented in \cref{eq:main} over the set $\mathcal{Q}$, which can be efficiently optimized with a few iterations as: 
\begin{equation}
\label{eq:codes}
\Q^* = \text{Diag}(\rr^{(t)}) \Q^{(0)} \text{Diag}(\cc^{(t)}).
\end{equation}
The renormalization vectors are computed using a small number of matrix multiplications via the iterative Sinkhorn-Knopp algorithm, where in each iteration:
\begin{align}
   & \rr^{(t)}=\kmarg/(\Q^{(0)} \cc^{(t-1)}), \label{renorm_1} \\
   & \cc^{(t)}=\nmarg/(\Q^{(0)} \rr^{(t)}),   \label{renorm_2}    
\end{align}
where $\cc^{(0)}=\mathbf{1}_{(N+M)}$. Also, $\Q^{(0)}$ is initialized as $\Q^{(0)}=(\exp(\mathbf{S}/\temp)/\sum(\exp(\mathbf{S}/\temp))$, with $\temp$ representing a temperature scaling parameter that controls the strength of the entropic constraint in \cref{eq:main}. Upon convergence, the matrix $\Q^*$ is normalized to produce soft class assignments that sum one for each sample, i.e., $\sum_k q_{ki}^*={\mathbf{1}_{(N+M)}\top}$.

\mypar{Producing conformal sets through codes.} Given the optimized matrix of codes, $\Q^*$, the final step consists of producing conformal sets from the obtained soft codes, $\vq_i^{*}$ (columns in $\Q^{*}$), for each query sample. To do so, calibration, $\{({\vq_i^{*}}^\top,y_i)\}_{i=1}^{N}$, and test, $\{({\vq_i^{*}}^\top)\}_{i=N+1}^{N+M}$, sets are separated again. Given an arbitrary non-conformity score, the predictive sets are created as detailed in \cref{background_conformal}: \textit{i}) generating non-conformity scores from codes for calibration data, $s_{i}=\mathcal{S}({\vq_i}^\top,y_i)$, \textit{ii}) finding the user-specified $1-\alpha$ quantile as in \cref{eq:threshold}, and \textit{iii}) creating conformal sets on test data based on such threshold following \cref{eq:inference}.

\begin{algorithm}
\caption{Conf-OT conformal prediction.}\label{alg:confot}
\begin{algorithmic}[1]
\State \textbf{input:} calibration dataset \domain{cal}$=\{(\vl_i,y_i)\}_{i=1}^{N}$, query set \domain{test}$=\{(\vl_i)\}_{i=N+1}^{N+M}$, non-conformity score function $\mathcal{S}$, error level $\alpha$, entropic weight $\temp$, iterations $T$.

\noindent \textcolor{gray}{// \textbf{Block 1.} - Transductive transfer learning.} \newline
\noindent \textcolor{gray}{// \textbf{Step 1.1.} - Init. optimal transport problem.}
\State $\mathbf{S}\in \real^{K\times(N+M)} = \left[l_{ki} \right]_{k=1,i=1}^{k=K,i=N+M}$ \textcolor{gray}{// Sim. matrix.}
\State $\kmarg=\frac{1}{N}\sum_1^N \yy_i$ \textcolor{gray}{// Label-marginal.}
\State $\nmarg = \frac{1}{(N+M)} \mathbf{1}_{(N+M)}$ \textcolor{gray}{// Sample marginal.}

\noindent \textcolor{gray}{// \textbf{Step 1.2.} - Compute renormalization vectors.}
\State $\Q^{(0)}=(\exp(\mathbf{S}/\temp)/\sum(\exp(\mathbf{S}/\temp))$ \textcolor{gray}{// Init. codes.}
\State $\cc^{(0)}=\mathbf{1}_{(N+M)}$ \textcolor{gray}{// Init. renormalization vector.}
\For{$t$ in $[1, \dots, T]$}
    \State $\rr^{(t)}=\kmarg/(\Q^{(0)} \cc^{(t-1)})$ \textcolor{gray}{// \cref{renorm_1}.}
    \State $\cc^{(t)}=\nmarg/(\Q^{(0)} \rr^{(t)})$ \textcolor{gray}{// \cref{renorm_2}.}
\EndFor

\noindent \textcolor{gray}{// \textbf{Step 1.3.} - Compute codes.}
\State $\Q^* = \text{Diag}(\rr^{(T)}) \Q^{(0)} \text{Diag}(\cc^{(T)})$ \textcolor{gray}{// Transport codes.}
\State $\Q^* = {\Q^*} \text{Diag}(1/\sum_{k}q^*_{ki})$ \textcolor{gray}{// Normalize.}

\noindent \textcolor{gray}{// \textbf{Block 2.} - Conformal prediction.}
\State \domain{cal}$=\{({\vq_i^*}^\top,y_i)\}_{i=1}^{N}$, \domain{test}$=\{({\vq_i^*}^\top)\}_{i=N+1}^{N+M}$ 

\noindent \textcolor{gray}{// \textbf{Step 2.1.} - $1-\alpha$ non-conformity score quantile.}
\State $\{s_{i}\}_{i=1}^{N}=\{\mathcal{S}({\vq_i^*}^\top,y_i)\}_{i=1}^{N}$ \textcolor{gray}{// Non-conformity scores.}
\State $\hat{s} \leftarrow \{s_{i}\}_{i=1}^{N}, \alpha$ \textcolor{gray}{// Search threshold - \cref{eq:threshold}.}

\noindent \textcolor{gray}{// \textbf{Step 2.2.} - Create query sets.}
\State \textbf{return:} $\{\mathcal{C}({\vq_i^*}^\top)\}_{i=N+1}^{M}$ \textcolor{gray}{// \cref{eq:inference}.}
\end{algorithmic}
\end{algorithm}

\mypar{Efficiency remarks.} Immediate concerns might arise regarding the computational feasibility of Conf-OT. On the contrary, it is highly efficient, especially compared to other transductive pipelines. First, it operates over black-box logits, and second, it is \textit{training-free}, requiring only a few iterations of the Sinkhorn algorithm with commodity resources. For the most extensive datasets, e.g., ImageNet ($K=1,000$ and $(N+M)=50,000$), the whole procedure requires only 1.1 seconds of additional overhead compared to its inductive counterpart (running on commodity CPUs). We extensively study its efficiency in \cref{main:subsection_results} and \appensecref{appendix_results_comp_efficiency}. Its robustness to different data ratios for calibration samples and query batch sizes is explored in \appensecref{appendix_results_data_efficiency}.

\section{Experiments}
\label{main:section_experiments}

\subsection{Setup}
\label{main:subsection_setup}

\mypar{Datasets.} In this work, we leverage CLIP's zero-shot capabilities to deploy a large-scale benchmark of conformal inference strategies across a wide corpus of 15 datasets. Note that the main body of literature on conformal inference in vision \cite{aps,raps,conftr,confts} is deployed on narrower scenarios using specialized models. In contrast, we use standard datasets for CLIP's zero- and few-shot adaptation, which gathers a heterogeneous number of tasks, from general objects to action recognition or fine-grained categories in specialized applications. These are: Imagenet \cite{deng2009imagenet}, ImageNet-A \cite{imagenet_a}, ImageNetV2 \cite{imagenetV2}, ImageNet-R \cite{imagenet_r}, ImageNet-Sketch \cite{imagenetSketch}, SUN397 \cite{sun397}, FGVCAircraft \cite{aircraft}, EuroSAT \cite{eurosat}, StanfordCars \cite{stanfordcars}, Food101 \cite{food101}, OxfordPets \cite{oxfordpets}, Flowers102 \cite{flowers102}, Caltech101 \cite{caltech}, DTD \cite{dtd}, and UCF101 \cite{ucf101}. We refer the reader to \appensecref{appendix_datasets} for specific details on each task. The corresponding test partition from each dataset is employed for our conformal inference experiments by producing disjoint calibration and testing subsets.

\mypar{Implementation details.} We use CLIP \cite{radford2021learning} pre-trained models, using different backbones: ResNet-50 and ResNet-101 \cite{resnet}, and ViT-B/32, ViT-B/16, and ViT-L/14 \cite{dosovitskiy2020vit}. Also, we experiment with MetaCLIP \cite{xu2024demystifying} ViT-B/16 and ViT-H/14 backbones. Unless otherwise indicated, ablation studies are performed with CLIP ViT-B/16. The text encoder from each model is used to produce class-wise prototypes for each downstream category by using standard templates and category names \cite{zhou2022coop,gao2021clip}, e.g., "\texttt{A photo of a [CLS]}.". Note that these templates are indicated in \appensecref{appendix_datasets}. Then, logits for each sample are produced by computing the temperature-scaled cosine similarity as formalized in \cref{background_zeroshot}. The hyper-parameters of Conf-OT are fixed for all tasks: the entropic weight is set to $\tau=1$, and the repetitions in the Sinkhorn algorithm are $T=3$.

\mypar{Baselines.} Note that we find no clear candidate for tackling the proposed scenario: training-free black-box transductive adaptation of CLIP models over the logit space. Hence, we adjusted prior transductive approaches to operate in the logit space. First, TIM \cite{tim} is leveraged as a general transductive framework based on information maximization. Concretely, a modified version to incorporate the label-marginal distribution obtained from the calibration set using a Kullback-Leibler (KL) divergence, coined $\text{TIM}_{\text{KL}(\widehat{\kmarg}||\kmarg)}$, is employed, as well as a version using a uniform prior, $\text{TIM}_{\text{KL}(\widehat{\kmarg}||\uu_K)}$. Second, we include the recently proposed TransCLIP \cite{zanella2024boosting}, a GMM-based clustering method specially designed for VLMs. These baselines are formally introduced in \appensecref{appendix_baselines}.

\mypar{Conformal prediction algorithms.} Three popular non-conformity scores for classification are assessed. In particular, we employ LAC \cite{lac}, and two adaptive approaches, APS \cite{aps}, and RAPS \cite{raps}, to generate prediction sets at error rates of $\alpha \in \{0.1, 0.05\}$. We set the hyper-parameters in RAPS to $k_{\text{reg}}=1$ and $\lambda=0.001$. These values provided stable performance in \cite{raps}. Even though the authors in \cite{raps} provide different strategies for automatically fixing these values for set size or adaptiveness optimization, we avoid using additional validation splits for hyper-parameter tuning in our experimental setting.

\mypar{Experimental protocol and metrics.} The test subset from the target datasets is partitioned into equally-sized calibration and testing, following the standard split strategy in \cite{raps}\footnote{Experiments using smaller data ratios are in \appensecref{appendix_results_data_efficiency}.}. All experiments are repeated 20 times using different random seeds. We include discriminative performance metrics, such as Top-1 accuracy, and figures of merit typically employed in conformal prediction settings. Concretely, we compute the standard coverage (“Cov.”) and average set size (“Size”), as well as class-conditioned coverage gap (“CCV”), which was recently proposed as a measure of adaptiveness \cite{ding2024class}. These are formalized in \appensecref{appendix_metrics}.

\subsection{Main results}
\label{main:subsection_results}

\mypar{Enhancing SoTA conformal predictors.} First, we compare the effect of Conf-OT with the base version of each non-conformity score using zero-shot predictions, i.e., no transfer learning. Results in \cref{main_results} demonstrate the advantages of the proposed transductive approach to enhance recent conformal inference strategies. \textit{Conf-OT provides consistent smaller set sizes for all conformal methods while maintaining the empirical coverage guarantees} for both $\alpha \in \{0.1, 0.05\}$. As a figure of its merit, set sizes consistently decrease in a relative ratio of nearly 20$\%$ compared to the base version. Also, class conditional coverage is consistently improved over 0.7 points when $\alpha=0.1$ and 0.3 points when $\alpha=0.05$. These results underscore the value of considering the structure of the unlabeled test samples during prediction to achieve better adaptability across many categories. Last, it is worth mentioning that the discriminative performance is enhanced notably by 2.6$\%$ for CLIP ResNet-50 and 2.9$\%$ for CLIP ViT-B/16. The positive performance of Conf-OT is also observed for additional CLIP and MetaCLIP encoders, whose results are provided in \appensecref{appendix_results_backbones}. Results per dataset are in \appensecref{appendix_results_datasets}.

\begin{table}[t!]
\setlength{\tabcolsep}{3.0pt}
\centering
\scriptsize
\begin{tabular}{lcccccccc}
\toprule
\multicolumn{1}{c}{\multirow{2}{*}{Method}} & & \multicolumn{3}{c}{$\alpha=0.10$} &  & \multicolumn{3}{c}{$\alpha=0.05$}   \\ \cmidrule{3-5}\cmidrule{7-9}  
\multicolumn{1}{c}{}      & Top-1$\uparrow$       & Cov   & Size$\downarrow$ & CCV$\downarrow$ & & Cov.   & Size$\downarrow$ & CCV$\downarrow$  \\
\midrule 
\multicolumn{9}{l}{\textbf{CLIP ResNet-50}} \\ \hdashline
LAC \cite{lac}                                    & 54.7                       & 0.900      & 10.77                       & 9.82                       &      & 0.950      & 19.22                       & 5.91 \\
\our \hspace{2mm}w/ Conf-OT           & \our \best{57.3}\imp{+2.6} & \our 0.900 & \our \best{8.61}\imp{-2.2}  & \our \best{9.15}\imp{-0.7} & \our & \our 0.951 & \our \best{15.53}\imp{-3.7} & \our \best{5.61}\imp{-0.3} \\
\cmidrule{1-9}  
APS \cite{aps}                                   & 54.7                       & 0.900      & 16.35                       & 8.36                       &      & 0.950      & 26.50                       & 5.34 \\
\our \hspace{2mm}w/ Conf-OT           & \our \best{57.3}\imp{+2.6} & \our 0.900 & \our \best{12.94}\imp{-3.4} & \our \best{7.64}\imp{-0.7} & \our & \our 0.950 & \our \best{20.96}\imp{-5.5} & \our \best{5.03}\imp{-0.3} \\
\cmidrule{1-9} 
RAPS \cite{raps}                                  & 54.7                       & 0.900      & 13.37                       & 8.46                       &      & 0.950      & 22.06                       & 5.44 \\
\our \hspace{2mm}w/ Conf-OT           & \our \best{57.3}\imp{+2.6} & \our 0.900 & \our \best{11.17}\imp{-2.2} & \our \best{7.72}\imp{-0.7} & \our & \our 0.950 & \our \best{17.24}\imp{-4.8} & \our \best{5.19}\imp{-0.3} \\
\midrule 
\multicolumn{9}{l}{\textbf{CLIP ViT-B/16}} \\ \hdashline 
LAC \cite{lac}                                   & 63.8                       & 0.899      & 5.52                       & 10.37                      &      & 0.950      & 10.24                      & 6.14 \\
\our \hspace{2mm}w/ Conf-OT           & \our \best{66.7}\imp{+2.9} & \our 0.900 & \our \best{4.40}\imp{-1.1} & \our \best{9.48}\imp{-0.9} & \our & \our 0.949 & \our \best{7.99}\imp{-2.3} & \our \best{5.80}\imp{-0.3} \\
\cmidrule{1-9}  
APS \cite{aps}                                   & 63.8                       & 0.900      & 9.87                       & 8.39                       &      & 0.950      & 16.92                        & 5.51 \\
\our \hspace{2mm}w/ Conf-OT           & \our \best{66.7}\imp{+2.9} & \our 0.899 & \our \best{7.64}\imp{-2.2} & \our \best{7.44}\imp{-1.0} & \our & \our 0.949 & \our \best{12.58}\imp{-4.3} & \our \best{5.09}\imp{-0.4} \\
\cmidrule{1-9} 
RAPS \cite{raps}                                 & 63.8                       & 0.900      & 8.12                       & 8.50                       &      & 0.950       & 12.66                        & 5.52 \\
\our \hspace{2mm}w/ Conf-OT          & \our \best{66.7}\imp{+2.9} & \our 0.900 & \our \best{6.68}\imp{-1.4} & \our \best{7.48}\imp{-1.0} & \our & \our 0.949  & \our \best{10.11}\imp{-2.6} & \our \best{5.16}\imp{-0.4} \\
\bottomrule
\end{tabular}
\caption{\textbf{Conf-OT performance} atop popular non-conformity scores, i.e., LAC \cite{lac}, APS \cite{aps}, and RAPS \cite{raps}. Average performance across 15 datasets. “$\downarrow$" indicates smaller values are better.}
\label{main_results}
\end{table}

\mypar{Comparison to transductive baselines.} Conf-OT is compared with relevant baselines in \cref{tab:transductive_baselines}. The evaluation extends not only to the performance but also to the computational efficiency of such methods. The latter is of special relevance in the explored application since base conformal inference methods do not produce considerable overhead during inference and are designed to operate in real-world scenarios with limited hardware resources. The figures of merit in \cref{tab:transductive_baselines} indicate that \textit{Conf-OT requires negligible additional inference times}. Also, \textit{Conf-OT requires no specific specialized hardware}, as it can run on commodity resources. In contrast, popular transductive methods require considerable GPU modules and inference times. While being a much more efficient solution, Conf-OT also excels in performance. For instance, TIM and TransCLIP deteriorate the produced set sizes when using LAC conformal score. Regarding adaptive scores such as APS and RAPS, all methods provide improvements over the base version, with $\text{TIM}_{\text{KL}(\widehat{\kmarg}||\uu_K)}$ outperforming Conf-OT when using APS. We explain this positive performance of TIM+APS by the effect of the entropy minimization term on producing higher-confidence predictions, which positively affects APS \cite{confts}. However, such a positive trend is a mirage that does not hold when evaluated across additional backbones and coverage rates, as shown in \appensecref{appendix_results_transduction}. Also, none of the considered methods improve the class-conditional coverage. Indeed, TransCLIP fails to provide the desired marginal coverage rate. Its Laplacian regularization term may cause this, as it is a neighborhood-based term that does not provide a joint optimization of calibration and test sets. The limitations of SoTA transductive methods enhance the qualities of the proposed solution. \textit{Conf-OT is a training-free solution that provides consistently smaller conformal sets compared to SoTA, maintaining coverage guarantees}. As an additional bonus, Conf-OT also provides the best result regarding discrimination, i.e., Top-1 accuracy.

\begin{table}[t!]
\setlength{\tabcolsep}{4pt}
\centering
\scriptsize
\begin{tabular}{lccccccc}
\toprule
\multicolumn{1}{c}{\multirow{1}{*}{Method}} & & & & \multicolumn{3}{c}{$\alpha=0.10$}\\ \cmidrule{5-8}
\multicolumn{1}{c}{}      & Top-1$\uparrow$  & T & GPU & Cov.   & Size$\downarrow$ & CCV$\downarrow$\\
\midrule 
LAC \cite{lac}                                                               & 63.8                        & \best{0.42} & -         & 0.899      & 5.52                       & 10.37                            \\ \hdashline
\hspace{1mm}$\text{TIM}_{\text{KL}(\widehat{\kmarg}||\uu_K)}$ \cite{tim}    & 64.7\imp{+0.9}              & 11.05       & 8.24      & 0.899      & 8.30\wor{+2.8}             & 10.41\wor{+0.0}                  \\
\hspace{1mm}$\text{TIM}_{\text{KL}(\widehat{\kmarg}||\kmarg)}$ \cite{tim}   & 65.0\imp{+1.2}              & 11.03       & 8.24      & 0.898      & 7.73\wor{+2.2}             & 10.89\wor{+0.5}                  \\
\hspace{1mm}TransCLIP \cite{zanella2024boosting}                            & 65.1\imp{+1.3}              & 12.00       & 12.2      & \nc{0.892} & 5.76\wor{+0.2}             & 11.02\wor{+0.7}                  \\
\our \hspace{1mm}Conf-OT                                                    & \our \best{66.7}\imp{+2.9} & \our 0.61   & \our -    & \our 0.900 & \our \best{4.40}\imp{-1.1} & \our \best{9.48}\imp{-0.9}       \\
\midrule 
APS \cite{aps}                                                              & 63.8                        & \best{0.54} & -         & 0.900      & 9.87                       & 8.39                             \\ \hdashline
\hspace{1mm}$\text{TIM}_{\text{KL}(\widehat{\kmarg}||\uu_K)}$ \cite{tim}    & 64.7\imp{+0.9}              & 11.16       & 8.24      & 0.900      & \best{7.24}\imp{-2.6}      & 9.32\wor{+0.9}                   \\
\hspace{1mm}$\text{TIM}_{\text{KL}(\widehat{\kmarg}||\kmarg)}$ \cite{tim}   & 65.0\imp{+1.2}              & 11.14       & 8.24      & 0.900      & 7.82\imp{-2.1}             & 9.38\wor{+1.0}                   \\
\hspace{1mm}TransCLIP \cite{zanella2024boosting}                            & 65.1\imp{+1.3}              & 12.12       & 12.2      & \nc{0.892} & 8.27\imp{-1.6}             & 11.50\wor{+3.1}                  \\
\our \hspace{1mm}Conf-OT                                                    & \our \best{66.7}\imp{+2.9}  & \our 0.72   & \our -    & \our 0.899 & \our 7.64\imp{-2.2}        & \our \best{7.44}\imp{-1.0}       \\
\midrule 
RAPS \cite{raps}                                                            & 63.8                        & \best{0.55} & -         & 0.900      & 8.12                       & 8.50                             \\ \hdashline
\hspace{1mm}$\text{TIM}_{\text{KL}(\widehat{\kmarg}||\uu_K)}$ \cite{tim}    & 64.7\imp{+0.9}              & 11.15       & 8.24      & 0.900      & 7.18\imp{-0.9}             & 9.32\wor{+0.8}                   \\
\hspace{1mm}$\text{TIM}_{\text{KL}(\widehat{\kmarg}||\kmarg)}$ \cite{tim}   & 65.0\imp{+1.2}              & 11.36       & 8.24      & 0.900      & 7.68\imp{-0.4}             & 9.42\wor{+0.9}                   \\
\hspace{1mm}TransCLIP \cite{zanella2024boosting}                            & 65.1\imp{+1.3}              & 12.12       & 12.3      & 0.899      & 7.17\imp{-1.0}             & 10.20\wor{+1.7}                  \\
\our \hspace{1mm}Conf-OT                                                    & \our \best{66.7}\imp{+2.9}  & \our 0.74   & \our -    & \our 0.900 & \our \best{6.68}\imp{-1.4} & \our  \best{7.48}\imp{-1.0}      \\
\bottomrule
\end{tabular}
\caption{\textbf{Comparison to transductive baselines}. Results using CLIP ViT-B/16 on 15 datasets. “T" refers to runtime in seconds, and “GPU" to peak memory usage (Gb).}
\label{tab:transductive_baselines}
\end{table}

\subsection{In-depth studies}
\label{main:subsection_ablation}

In the following, we provide additional experiments to explore the conformal inference on VLMs in a more detailed manner, as well as key features of the proposed Conf-OT.

\begin{figure*}[t!]
    \begin{center}
        \begin{tabular}{cccc}

         \includegraphics[width=.25\linewidth]{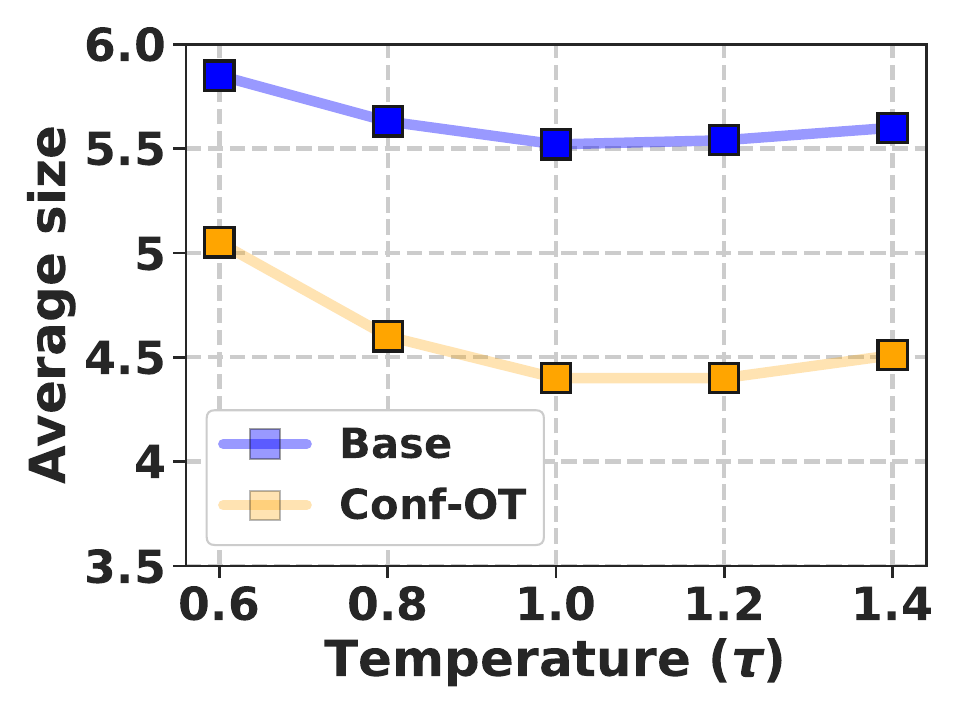} &
         \includegraphics[width=.25\linewidth]{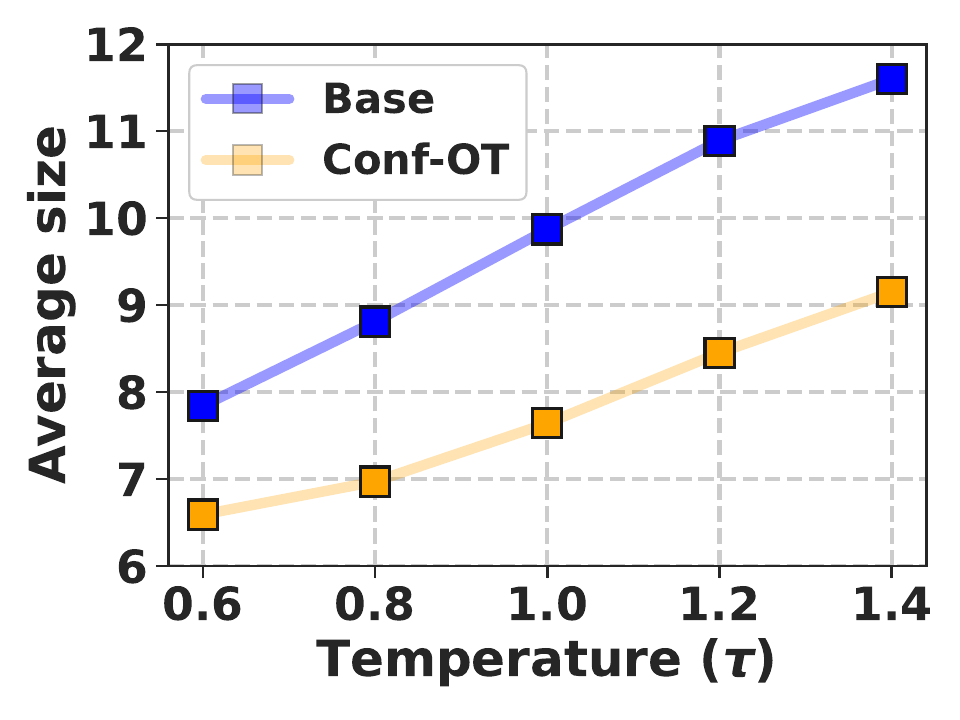} &
         \includegraphics[width=.25\linewidth]{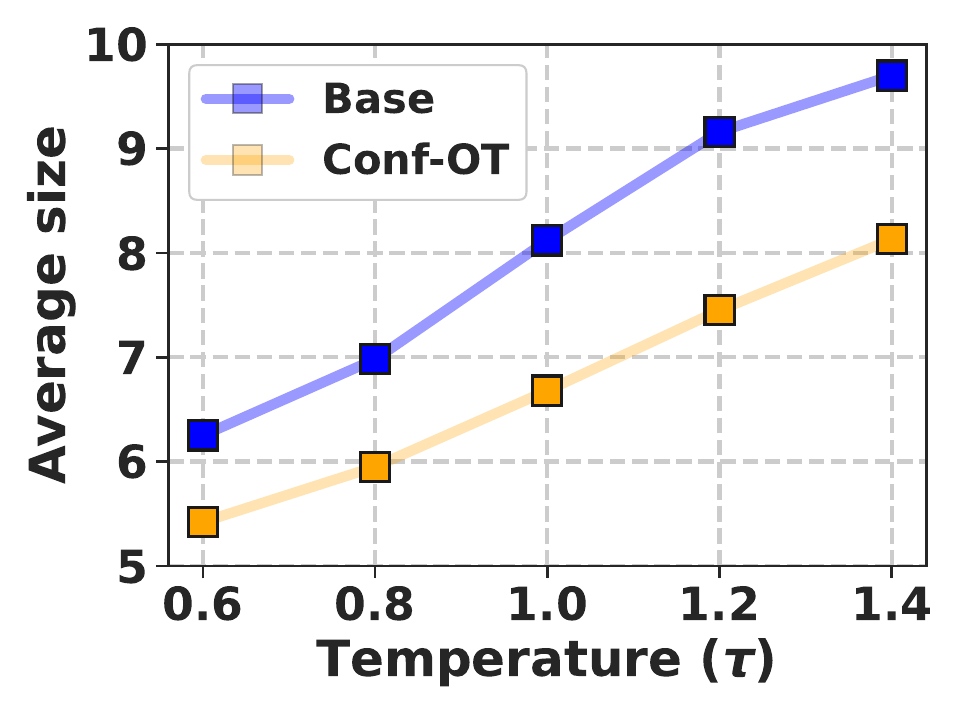} & \\
        (a) LAC \cite{lac}   & (b) APS \cite{aps} & (c) RAPS \cite{raps} & \\

        \end{tabular}
        \caption{\textbf{Entropic constraint ($\tau$).} Conf-OT is compatible with recent observations \cite{confts} regarding the positive effect on set size of temperature scaling ($\tau<1$) on adaptive scores (b,c). However, such behavior does not generalize to non-adaptive scores, i.e., LAC (a), whereas Conf-OT improves the performance atop all non-conformity scores. Results using CLIP ViT-B/16 on 15 datasets with $\alpha=0.10$.}
        \label{fig:distributions}
    \end{center}
\end{figure*}

\mypar{Complementary to temperature scaling.} Conformal inference is usually related to other uncertainty frameworks, such as calibration. Notably, previous literature \cite{raps} tends to integrate calibration steps such as temperature scaling (TS). Recently, the authors in  \cite{confts} explored the impact of temperature scaling on adaptive scores (APS, RAPS), observing that high-confidence predictions ($\tau<1$) lead to smaller sets on average. The Sinkhorn optimal transport solver incorporates entropic constraints through a temperature-scaling parameter (see \cref{alg:confot}), potentially affecting the conformal sets. Hence, we consider this aspect to deserve a specific study. \cref{fig:distributions} illustrates the joint effect of TS and Conf-OT. Results follow the observations in \cite{confts}: TS with $\tau<1$ improves efficiency in adaptive methods, also when combined with Conf-OT. Notably, Conf-OT improvements are orthogonal to the ones produced by simply the sharpness of the probability distribution since it also controls other aspects, e.g., the label-marginal distribution in the overall assignment. Thus, while the effect of TS using non-adaptive scores such as LAC is absent, Conf-OT consistently improves. Based on these observations, we kept $\tau=1$ for the entropic constraint weight in Conf-OT to provide a general framework for all non-conformity scores.

\begin{figure}[t!]
\setlength{\tabcolsep}{1.0pt}
    \begin{center}
        \begin{tabular}{cc}

         \hspace{-2mm}
         \includegraphics[width=.50\linewidth]{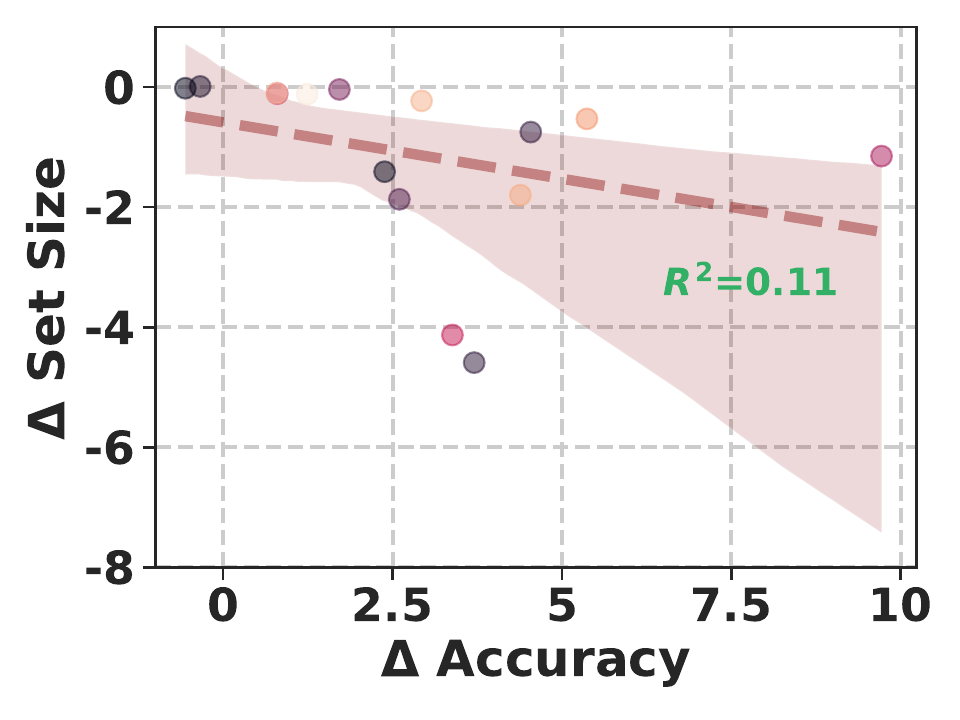} &
         \includegraphics[width=.50\linewidth]{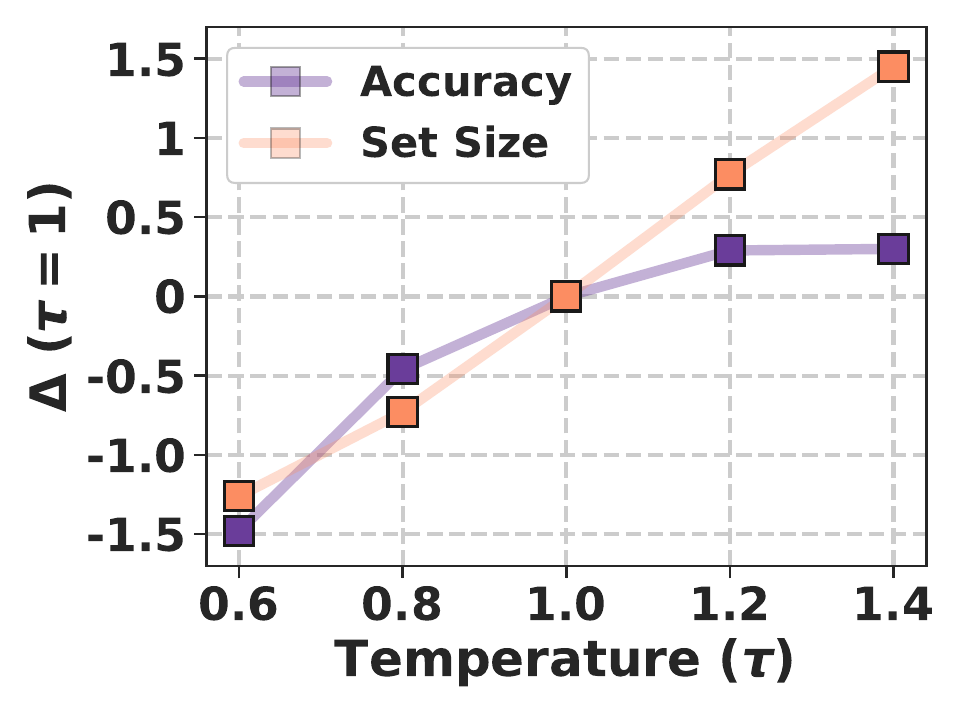} \\
        (a) & (b) \\

        \end{tabular}
        \caption{\textbf{Accuracy \textit{vs.} set size change ($\Delta$) using Conf-OT}. (a) Correlation among datasets for LAC \cite{lac}. (b) Effect of the entropic constraint for RAPS \cite{raps}. Results using CLIP ViT-B/16 on 15 datasets with $\alpha=0.10$. More information in \appensecref{appendix_results_set_accuracy}.}
        \label{fig:accuracy_vs_size}
    \end{center}
\end{figure}

\mypar{Is the improvement \textit{only} on the discriminative aspect?} One could argue that the better behavior of Conf-OT is explained by producing the largest probabilities on the correct class more often, i.e., discriminative performance, as observed in Top-1 accuracies improvement in \cref{main_results}. However, we have observed that this is not the case. For example, \cref{fig:accuracy_vs_size}(a) shows a limited correlation between size and accuracy across datasets for LAC. Also, \cref{fig:accuracy_vs_size}(b) illustrates an inverse trend when further optimizing the entropic constraint in adaptive conformal methods. Additionally, although all transductive baselines improve accuracy, they sometimes do not produce better conformal sets in \cref{tab:transductive_baselines}. These observations indicate a disjoint behavior between discriminative and conformal inference figures. 

\mypar{Conf-OT components.} The proposed approach presents a small number of tunable elements. First, as previously discussed, we fixed the entropic constraint weight to its standard value, $\tau=1$. Second, we fixed the number of repetitions in the Sinkhorn algorithm to 3. These are enough for convergence, as shown in \appensecref{appendix_results_hyperparam}. Last, Conf-OT uses the label-marginal distribution of the calibration set to constrain the optimal transport problem. \cref{tab:ablation_marginal} provides figures that showcase the importance of this element. It is worth mentioning that the potential of accessing this marginal distribution in transductive settings is not new. Indeed, oracle scenarios in image segmentation have also pointed in this direction \cite{repri}. Nevertheless, the standard conformal inference setting grants access to this information to ensure the assumption of data exchangeability \cite{vovk_book}. The constraints of the Sinkhorn algorithm excel at efficiently incorporating such priors, especially compared to the other resource-demanding transduction baselines.

\begin{table}[t!]
\centering
\scriptsize
\begin{tabular}{llccccc}
\toprule
\multicolumn{1}{c}{\multirow{1}{*}{Method}} & \multicolumn{1}{c}{\multirow{1}{*}{Prior}}  & & \multicolumn{3}{c}{$\alpha$ = 0.10}\\ \cmidrule{4-6}
&  \multicolumn{1}{c}{}      & Top-1$\uparrow$       & Cov.   & Size$\downarrow$ & CCV$\downarrow$  \\
\midrule 
LAC \cite{lac}                                               &               & 63.8                       & 0.899      & 5.52                       & 10.37                      \\ \hdashline
\multicolumn{1}{l}{\multirow{1}{*}{\hspace{2mm}w/ Conf-OT}} &  $\kmarg = \uu_K$      & 65.5\imp{+1.7}             & 0.900      & 5.32\imp{-0.2}             & 9.87\imp{-0.5}             \\
                                               \our \hspace{2mm}w/ Conf-OT  & \our \textit{Ours}    & \our \best{66.7}\imp{+2.9} & \our 0.900 & \our \best{4.40}\imp{-1.1} & \our \best{9.48}\imp{-0.9} \\
\midrule 
APS \cite{aps}                                              &               & 63.8                       & 0.900      & 9.87                       & 8.39                       \\ \hdashline
\multicolumn{1}{l}{\multirow{1}{*}{\hspace{2mm}w/ Conf-OT}}  &  $\kmarg = \uu_K$      & 65.5\imp{+1.7}             & 0.900      & 8.72\imp{-1.2}             & \best{7.31}\imp{-1.1}      \\
                                                 \our \hspace{2mm}w/ Conf-OT & \our \textit{Ours}    & \our \best{66.7}\imp{+2.9} & \our 0.899 & \our \best{7.64}\imp{-2.2} & \our 7.44\imp{-1.0}        \\
\midrule 
RAPS \cite{raps}                                             &               & 63.8                       & 0.900      & 8.12                       & 8.50                       \\ \hdashline
\multicolumn{1}{l}{\multirow{1}{*}{\hspace{2mm}w/ Conf-OT}} &  $\kmarg = \uu_K$      & 65.5\imp{+1.7}             & 0.900      & 7.57\imp{-0.6}             & \best{7.31}\imp{-1.2}      \\
                                              \our \hspace{2mm}w/ Conf-OT    & \our \textit{Ours}    & \our \best{66.7}\imp{+2.9} & \our 0.900 & \our \best{6.68}\imp{-1.4} & \our 7.48\imp{-1.0}        \\
\bottomrule
\end{tabular}
\caption{\textbf{Role of label-marginal prior} ($\kmarg$ in \cref{polytope}) in Conf-OT. Results using CLIP ViT-B/16 and averaged over 15 datasets.}
\label{tab:ablation_marginal}
\end{table}

\mypar{Data efficiency.} We delve into this aspect of our transductive strategy in two measures: the calibration data ratio and robustness to small query sets. Specific numbers are in \appensecref{appendix_results_data_efficiency}. These demonstrate that \textit{the efficiency of the sets produced by Conf-OT holds even in the most challenging scenarios}, e.g., using only 10$\%$ of data for calibration or receiving extremely small query sets of 8 or 16 images.

\section{Limitations}
\label{main:section_limitations}

In this work, we have explored the conformal prediction framework for zero-shot VLMs. To alleviate the absence of an adaptation stage, we have introduced a transductive setting to enhance the efficiency and adaptiveness of any conformal score by leveraging well-established knowledge in optimal transport. Our method is effective, but it presents some limitations. These are inherited from its transductive and conformal prediction nature. Particularly, it is valid under data exchangeability assumptions to guarantee the desired coverage, like any other conformal prediction method, and requires additional resources during inference, yet being $\times$15 faster than other transductive approaches.

\section*{Acknowledgments}
This work was funded by the Natural Sciences and Engineering Research Council of Canada (NSERC). We also thank Calcul Québec and Compute Canada.

{\small
\bibliographystyle{ieeenat_fullname}
\bibliography{refs}
}

\appendix

\setcounter{section}{0}
\renewcommand{\thesection}{\Alph{section}}

\maketitlesupplementary

\section{Extended related works}
\label{rel_work_extended}

In the following, we provide extended remarks about prior literature on conformal prediction. Concretely, we delve into its historical transductive nature and two different families of methods for further improving the conformal inference in vision: conformal training and methods that assume access to additional data splits.

\mypar{Transduction in conformal prediction.} Initially, this reliability framework was transductive \cite{learning_ny_transduction,transductive_ci,trans_conf_machine}. The original setting, usually called \textit{full conformal prediction}, assumes access to \textit{\iid} labeled training and unlabeled test samples for producing the conformal sets. These methods require evaluating all the label space, $\mathcal{Y}=\{1, 2, ..., K\}$, to fit label-specific models, including all values of $y$ that are sufficiently consistent for a given significance level. Such computational overload prohibits its use in modern deep-learning models. In contrast, \textit{split conformal prediction} \cite{inductive_ci,vovk_book} assumes a more resource-efficient, practical setting. Given a trained, black-box model that outputs logit predictions, it assumes access to a fresh labeled calibration set exchangeable \cite{vovk_book} to testing data. Even though it was initially known as \textit{inductive conformal prediction} \cite{inductive_ci,pmlr-v25-vovk12}, split conformal prediction does not necessarily preclude transduction mechanisms \cite{2024transconf}. Our transductive approach differs from these initial conceptions on full conformal prediction but exploits transduction while adapting a black-box model following the split conformal prediction setting.

\mypar{Conformal training in vision.} A family of works, known as the \textit{conformal training}, focuses on preparing the base model to enforce small predictive sets \cite{conftr,conftr_24} or improving sample adaptiveness by regularizing learning to produce homogenous non-conformity scores \cite{einbinder2022training}. Conformal training methods \cite{conftr,conftr_24,einbinder2022training} are not applicable in the era of foundation models, where models are trained on limited occasions using general learning objectives and accessed for efficient adaptation. Thus, they do not apply to the explored black-box setting and fall out of the scope of this paper.

\mypar{Improving conformal sets with additional data splits.} Within the black-box split conformal prediction, recent literature explores novel aspects of well-known non-conformity scores, such as APS or RAPS. For example, \cite{conftr} explores the impact of temperature scaling, and \cite{ding2024class} studies its behavior on tasks with many classes and, more particularly, the class-conditional coverage these scores offer. Also, the authors propose different strategies for improving these approaches. First, Conf-OT \cite{conftr} proposes to train the optimum temperature scaling by minimizing the efficiency gap in a costume loss function. Also, Clustered Conformal Prediction \cite{ding2024class} integrates a clustering step
to find subgroups of classes with similar quantized behavior for a given non-conformity score using K-means. The groups are then employed to perform conformal prediction at the cluster level. Such measures have shown improvement over the base adaptive methods, i.e., APS and RAPS. However, this comes at the cost of incorporating additional data splits to adjust the proposed methods without potentially breaking the exchangeability to test data. In this work, we stick to the standard experimental setting by accessing one calibration set uniquely. We argue that this setting is more realistic, especially in critical scenarios such as detecting rare, low-prevalence diseases \cite{fundusrare,skinrare,shakeri2024few}.

\begin{table*}[!ht]
\setlength{\tabcolsep}{3.0pt}
\centering
\scriptsize
\begin{tabular}{lrrrrcll}
\toprule
\multicolumn{1}{c}{Dataset} & Classes & \multicolumn{3}{c}{Splits} & b/u & \multicolumn{1}{c}{Task description} & \multicolumn{1}{c}{Text templates}\\
                            & & \multicolumn{1}{c}{Train} & \multicolumn{1}{c}{Val} & \multicolumn{1}{c}{Test} & & & \\
\midrule
ImageNet \cite{deng2009imagenet}        & 1,000 & 1.28M & - & 50,000 & $b$  & Natural objects recognition.               & \multirow{5}{*}{\begin{tabular}[l]{@{}l@{}} "\texttt{Itap of a [CLS]}." , "\texttt{A bad photo of [CLS]}." , \\  "\texttt{A origami of [CLS]}." , "\texttt{A photo of the large [CLS]}." , \\ "\texttt{A [CLS] on a video game}." , "\texttt{Art of the [CLS]}." , \\ "\texttt{A photo of the small [CLS]}." , "\texttt{A photo of a [CLS]}." \end{tabular}} \\
ImageNet-A \cite{imagenet_a}            & 200   & -      & -      & 7,500   & $u$ & Natural objects recognition.               &  \\
ImageNet-V2 \cite{imagenetV2}           & 1,000 & -      & -      & 10,000  & $b$ & Natural objects recognition.               &  \\
ImageNet-R \cite{imagenet_r}            & 200   & -      & -      & 30,000  & $u$ & Natural objects recognition.               &  \\
ImageNet-Sketch \cite{imagenetSketch}   & 1,000 & -      & -      & 50,889  & $b$ & Sketch-style images.         &  \\
\midrule
SUN397 \cite{sun397}                    & 397   & 15,880 & 3,970  & 19,850  & $b$ & Scenes classification.                   & "\texttt{A photo of a [CLS]}." \\
FGVCAircraft \cite{aircraft}            & 100   & 3,334  & 3,333  & 3,333   & $u$ & Aircraft classification.                 & "\texttt{A photo of [CLS], a type of aircraft}." \\
EuroSAT \cite{eurosat}                  & 10    & 13,500 & 5,400  & 8,100   & $u$ & Satellite image classification.          & "\texttt{A centered satellite photo of [CLS]}." \\
StanfordCars \cite{stanfordcars}        & 196   & 6,509  & 1,635  & 8,041   & $u$ & Cars classification.                     & "\texttt{A photo of a [CLS]}." \\
Food101 \cite{food101}                  & 101   & 50,500 & 20,200 & 30,300  & $b$ & Foods classification.                    & "\texttt{A photo of a [CLS], a type of food}." \\
OxfordPets \cite{oxfordpets}            & 37    & 2,944  & 736    & 3,669   & $u$ & Pets classification.                     & "\texttt{A photo of a [CLS], a type of a pet}." \\
Flowers102 \cite{flowers102}            & 102   & 4,093  & 1,633  & 2,463   & $u$ & Flowers classification.                  & "\texttt{A photo of a [CLS], a type of flower}."\\
Caltech101 \cite{caltech}               & 100   & 4,128  & 1,649  & 2,465   & $u$ & Natural objects classification.          & "\texttt{A photo of a [CLS]}." \\
DTD \cite{dtd}                          & 47    & 2,820  & 1,128  & 1,692   & $b$ & Textures classification.                 & "\texttt{[CLS] texture}." \\
UCF101 \cite{ucf101}                    & 101   & 7,639  & 1,898  & 3,783   & $u$ & Action recognition.                      & "\texttt{A photo of a person doing [CLS]}." \\
\bottomrule
\end{tabular}
\caption{\textbf{Datasets overview}. Detailed description of the 15 datasets used to evaluate the conformal inference of zero-shot vision-language models. Also, the handcrafted textual templates for setting the zero-shot text-driven classifier for each dataset are indicated. These are the same ones used in relevant prior literature on this topic \cite{zhou2022coop,clap24}. “$b$"/“$u$" denotes balanced or unbalanced test partitions, respectively.}
\label{datasets}
\end{table*}

\section{Non-conformity scores}
\label{appendix_scores}

The following formally introduces the non-conformity scores employed in this paper. The measures are designed so that smaller conformal scores correspond to larger model confidence levels. These are employed to search its quantile that satisfies the 1-$\alpha$ coverage in the calibration set, following \cref{eq:threshold}, and producing conformal sets in test samples as detailed in \cref{eq:inference} in the main manuscript.

\mypar{Least Ambiguos Classifier.} The intuitive idea of LAC \cite{lac} is constructing sets by thresholding the output probabilities with a confidence level. Thus, the non-conformity score can be constructed as:
\begin{align}
\label{eq:lac}
    \mathcal{S}_{\text{ LAC}}(\xx,y) = 1-x_{k=y}.
\end{align}

LAC produces the minimum set sizes in case the input probabilities are correct. However, it lacks adaptiveness, e.g., in under-represented categories.

\mypar{Adaptive Prediction Sets.} Aiming to improve adaptiveness, APS \cite{aps} construct the confidence sets based on accumulating ordered class confidences, such that:
\begin{align}
\label{eq:aps}
    \mathcal{S}_{\text{ APS}}(\xx,y) = \rho_x(y) + x_{k=y} \cdot u,
\end{align}
\noindent where $\rho_x(y)$ is the total probability mass of the set of labels more likely than the input label $y$, i.e., $\rho_x(y)=\sum_{k'\in\mathcal{Y'}(\xx,y)} x_{k=k'}$, with $\mathcal{Y'}(\xx,y)=\{k|x_{k}>x_{k=y}\}$. Also, note that $u\in[0,1]$ is a random variable to break ties and archive exact 1-$\alpha$ marginal coverage.

\mypar{Regularized Adaptive Prediction Sets.} Even though APS produces better coverage than LAC across different data subgroups, it comes at the cost of producing large set sizes. To tackle this issue, RAPS \cite{raps} incorporates penalties when adding categories into the accumulative confidence procedure, thus taming the score distribution tail. \\ RAPS can be formally introduced as:
\begin{align}
\label{eq:raps}
    \mathcal{S}_{\text{ RAPS}}(\xx,y) = \mathcal{S}_{\text{ APS}}(\xx,y) + \lambda \cdot (o(\xx,y)-k_{\text{reg}})^{+},
\end{align}
\noindent where $\lambda,k_{\text{reg}}\geq0$ are hyper-parameters that control the strength of the constraint, $o_x(y)$ is the rank of the label y in the sorted predictions, $o(\xx,y)=|\mathcal{Y'}(\xx,y)|+1$, and $(\cdot)^+$ represented the ReLU function. Regarding the hyper-parameters, $\lambda$ is the additional confidence cost of incorporating each new category to the set, and $k_{\text{reg}}$ is the category index in which the penalty starts.

\section{Additional datasets details}
\label{appendix_datasets}

\mypar{Datasets}. As stated in the main manuscript, we perform a large-scale benchmark on the conformal prediction of CLIP models across typical datasets employed for transferability \cite{radford2021learning,gao2021clip,zhou2022coop}. Concretely, we employ 15 datasets, which compile thousands of general concepts, fine-graned categories, textures, and actions. Also, it is worth noting that these benchmarks include 9/15 unbalanced tasks, which are commonly more challenging to address in vision classification literature. A summary of the employed datasets and task descriptions is depicted in \cref{datasets}. Regarding the dataset splits, we employed the ones proposed in seminal works for few-shot adaptation of CLIP \cite{gao2021clip,zhou2022coop}.

\mypar{CLIP's text templates}. For creating text prototypes for zero-shot and transfer learning using CLIP, the target category names ("\texttt{[CLS]}.") are forwarded through the text encoder. These category names are usually combined with hand-crafted text templates, which provide additional context on the task at hand, \,  e.g., "\texttt{A centered satellite photo of [CLS]}." We followed prior works in this aspect \cite{zhou2022coop,clap24}, using common text templates for each task, which are depicted in \cref{datasets}.

\section{Evaluation of conformal inference}
\label{appendix_metrics}

The evaluation metrics indicated in the experimental section are formally introduced in this section.

\mypar{Top-1 accuracy.} To evaluate the discriminative performance of CLIP models, we employ accuracy, a widely extended metric in few-shot literature \cite{gao2021clip}, and conformal prediction in vision \cite{raps}.

The following details the metrics employed to evaluate the conformal inference methods. For that purpose, let us assume an arbitrary data set $\mathcal{D}=\{(\xx_i,y_i)\}_{i=1}^{I}$, an error rate of coverage level $\alpha$, and the function creating the conformal sets from a non-conformity score $C(\xx)$, which operates as in \cref{eq:inference}, after finding the non-conformity score threshold from a calibration subset. We refer the reader to \cref{background_conformal} in the main manuscript for other definitions.

\mypar{Coverage.} The empirical coverage on the test domain is employed to measure the degree of satisfaction of the marginal coverage guarantees:
\begin{equation}
    \label{eq:cov}
    \text{Cov}(\mathcal{D})=\frac{1}{I} \sum_{i\in\mathcal{D}} \delta[(y_i \subset C(\xx_i)],
\end{equation}

\noindent where $\delta$ denotes a delta function, that is, 1 if its argument is true, and 0 otherwise.

\mypar{Set size.} The average set size, also known as inefficiency \cite{conftr}, is a widely employed metric \cite{raps,ding2024class,confts} for assessing the utility of conformal prediction methods for multi-class classification problems. An optimum conformal method in terms of efficiency should provide a lower set size:
\begin{equation}
    \label{eq:set_size}
    \text{Size}(\mathcal{D})=\frac{1}{I} \sum_{i\in\mathcal{D}} |C(\xx_i)|.
\end{equation}

\mypar{Class-conditional coverage violation (CCV).} A recently proposed metric in \cite{ding2024class} to evaluate the adaptiveness of a conformal prediction method based on measuring the empirical coverage gap observed in each category in the target task. An optimal conformal method in terms of adaptiveness is expected to provide a small average gap:
\begin{equation}
    \label{eq:covgap}
    \text{CCV}(\mathcal{D})=100 \times \frac{1}{|\mathcal Y |}\sum_{k\in \mathcal Y} \big\vert \text{Cov}(\mathcal D_{k}) - (1-\alpha)\big\vert,
\end{equation}

\noindent where $|\cdot|$ over the scalar in the summation term represents the absolute value, and $\mathcal D_{k}$ indicates the subset of samples labeled as the category $k$, such that $\mathcal D_{k} = \{(\xx_i,y_i)\}_{i\in \mathcal{B}_k}$, with $\mathcal{B}_k = \{i \ | \ y_i=k\}$. Note that the metric is multiplied by 100 to provide a percentage scale.

\section{Details on inductive adaptation}
\label{appendix_inductive}

\cref{fig:inductive}(b) refers to an experiment using the calibration set for adapting the zero-shot logits to the new tasks, so-called Adapt+SCP. Concretely, we train new class prototypes on the logit space, $\mathbf{W}\in \real^{K\times K}$. These weights are initialized in the simplex corners and are l2-normalized during training such that $\mathbf{w}_{k}=\frac{\mathbf{w}_{k}}{||\mathbf{w}_{k}||_2}$. Given the labeled calibration set, \domain{cal}$=\{(\vl_i,y_i)\}_{i=1}^{N}$, we first l2-normalize the received logit representations, $\vl=\frac{\vl}{||\vl||_2}$. Then, the new scores based on the learned class prototypes are obtained as follows:
\begin{align}
\label{eq:tim_logits}
    l_{k}' = - \frac{\tau^{\text{\tiny{LP}}}}{2} ||\vl-\mathbf{w}_{k}||,
\end{align}
\noindent where $\tau^{\text{\tiny{LP}}}$ is a temperature scaling parameter, which is searched greedily based on calibration data for each dataset, and $\vl'=(l_k')_{1 \leq k \leq K}$ are the new logits for a given sample. Finally, for a given sample, the new output probabilities are the softmax of the new logits: $\pp=\sigma_{k}(\vl')$.

The inductive adaptation consists of learning the new class weights based on a few shots available on calibration data by minimizing cross-entropy:
\begin{align}
\label{eq:tim_ce}
    \phantom{.}\min_{\mathbf{W}} \; - \frac{1}{NK} \sum_{i=1}^I \sum_{k=1}^K {y}_{ik} \log p_{ik},
\end{align}
\noindent where $\{\{{y}_{ik}\}_{i=1}^{I}\}_{k=1}^{K}$ are the one-hot-encoded labels.
Training is performed via mini-batch gradient descent, using large batches of $2,048$ samples and ADAM as an optimizer, with a learning rate of $0.1$, during $500$ iterations, using a cosine decay scheduler for the learning rate.

\section{Transductive adaptation baselines}
\label{appendix_baselines}

As stated in the main manuscript, there is no clear candidate to include as a baseline for the proposed setting: training-free, transductive adaptation of VLMs using black-box logit predictions. Hence, we adapted two transductive training approaches for the task: a general transductive formulation based on mutual information, i.e., TIM \cite{tim}, and the recently proposed TransCLIP \cite{zanella2024boosting}, a GMM-based method specially designed for zero-shot VLMs.

\mypar{Transductive information maximization \cite{tim}.} TIM is a general framework based on mutual information for adjusting a set of class prototypes on unlabeled data. Formally, the employed input logits, class weights, and new probabilities are obtained as in the inductive setting detailed in Section \appensecref{appendix_inductive}. More concretely, softmax probabilities are obtained following \cref{eq:tim_logits}. In contrast, TIM operates unsupervised and targets an entropy minimization loss with the regularized label-marginal distribution. We follow the SGD-based version proposed by the authors and modify the Shannon entropy maximization term by a KL divergence, which allows us to employ the observed label-marginal on the calibration set, i.e., $\kmarg$. Two versions are proposed, with different regularizations for the predicted label-marginal distribution, $\hat{\kmarg}=\frac{1}{N+M}\sum_{i=1}^{N+M} \pp_i$. These two versions are:

\noindent $\text{TIM}_{\text{KL}(\widehat{\kmarg}||\uu_K)}$: the base version, closer to the original work in \cite{tim} employes a uniform target label-marginal distribution, such that:
\begin{align}
\label{eq:tim_uniform}
    \phantom{.}\min_{\mathbf{W}} \; \frac{\alpha}{N} \sum_{i=1}^I \sum_{k=1}^K p_{ik} \log p_{ik} + \sum_{k=1}^K \uu_K \log \frac{\uu_K}{\hat{\kmarg}} \, .
\end{align}

\noindent $\text{TIM}_{\text{KL}(\widehat{\kmarg}||\kmarg)}$: the modified version, leverages the marginal distributions observed in the calibration set, whose annotated labels are available:
\begin{align}
\label{eq:tim_kl}
    \phantom{.}\min_{\mathbf{W}} \; \frac{\alpha}{N} \sum_{i=1}^I \sum_{k=1}^K p_{ik} \log p_{ik} + \sum_{k=1}^K \kmarg \log \frac{\kmarg}{\hat{\kmarg}} \, .
\end{align}

The training is performed by gradient descent, using large batches of $2,048$ samples during $100$ iterations, ADAM as an optimizer, and a base learning rate of $0.001$. The later hyper-parameters follow the advice provided in SGD-TIM \cite{tim}. Finally, it is worth mentioning that we found TIM highly sensitive to the choice of $\tau^{\text{\tiny{LP}}}$ in \cref{eq:tim_logits}. To alleviate this issue, $\tau^{\text{\tiny{LP}}}$ is searched using a grid of $\tau^{\text{\tiny{LP}}} \in \{0.1, 1, 5, 10, 15, 30, 60, 100\}$ per dataset. We employed the accuracy on the calibration set after transductive adaptation as the maximization objective for such a search. Note that these labels are available in the split conformal inference scenario, and we did not observe any empirical degradation in marginal coverage. Also, $\alpha$ is fixed to $\alpha=0.1$. 

\mypar{TransCLIP \cite{zanella2024boosting}.} We perform minor modifications over the base zero-shot version proposed by the authors. First, we set the zero-shot CLIP prototypes as the corners of the logit simplex. Second, we find the method relatively sensitive to the text-guided KL divergence penalty $\lambda$, which we increased to $\lambda=10$ upon a greedy search to avoid performance degradation. In terms of performance, our results obtained with this baseline are close to the figures reported by the authors in TransCLIP. Concretely, for ViT-B/16, we obtained an average accuracy of $65.1$ for 15 datasets. For the same tasks, the authors report $68.1$ accuracy. This performance gap is explained by the difference in the input given to TransCLIP since logit scores can be seen as using projections of the original features, with the consequent information loss, especially for tasks with small categories, such as EuroSAT. Concretely, this dataset mainly contributes to such an average accuracy gap ($-16.8$).

As a final remark, we would want to highlight the presence of key hyper-parameters on the explored baselines. The performance for each specific task might be sensitive to the choice of these. In contrast, \textit{Conf-OT does not introduce any critical hyper-parameter that could degrade the transfer learning performance.}

\section{Additional results}
\label{appendix_results}

This section introduces additional results and specific metrics that showcase Conf-OT's effectiveness compared to the prior art and validate its key elements and robustness across various scenarios. 

\subsection{Additional CLIP backbones}
\label{appendix_results_backbones}

In \cref{supp_results_backbones}, we report the performance of Conf-OT atop additional CLIP, i.e., ResNet-101, ViT-B/32, and ViT-L/14, and MetaCLIP, i.e., ViT-B/16, and ViT-H/14, backbones. The results are consistent with the main findings in \cref{main:subsection_results} (\cref{main_results}) and indicate the generalization of the proposed setting across several black-box outputs. Concretely, efficiency improvements on set size are consistently maintained on CLIP ResNet-101 and CLIP ViT-B/32, with such a figure improved by nearly $20\%$. Also, Conf-OT is effective on larger backbones such as CLIP ViT-L/14 or MetaCLIP ViT-H/14, whose initial performance is also notably better. In this case, relative performance improvements of nearly $15\%$ are observed.

\subsection{Results per dataset}
\label{appendix_results_datasets}

Performances are detailed for each backbone individually. Also, we split the five ImageNet shifts and 10 fine-grained tasks into different Tables to improve readability. First, \cref{appendix_results_rn50_10tasks} and \ref{appendix_results_rn50_imagenet} introduce the results for CLIP ResNet-50, and \cref{appendix_results_rn101_10tasks} and \ref{appendix_results_rn101_imagenet} do the same for CLIP ResNet-101. \cref{appendix_results_vit32_10tasks}, \ref{appendix_results_vitb32_imagenet}, and \ref{appendix_results_vit16_10tasks}, \ref{appendix_results_vitb16_imagenet} introduce the results for both CLIP ViT-B backbones: ViT-B/32 and ViT-B/16, respectively. Finally, \cref{appendix_results_vitl14_10tasks} and \ref{appendix_results_vitbl14_imagenet} present the results for ViT-L/14. Complementary, \cref{appendix_results_MetaCLIP_vitB16_10tasks}, \ref{appendix_results_MetaCLIP_vitB16_imagenet}, \ref{appendix_results_MetaCLIP_vith14_10tasks}, and \ref{appendix_results_MetaCLIP_vith14_imagenet} depict the detailed results for MetaCLIP backbones across all datasets.

\subsection{Additional results for transductive baselines}
\label{appendix_results_transduction}

\cref{tab:supp_baselines} introduces the performance obtained using TIM \cite{tim} and TransCLIP \cite{zanella2024boosting} baselines for additional datasets, i.e., CLIP ResNet-50, and more demanding error rates, i.e., $\alpha=0.05$. These results complement the observations in \cref{main:subsection_results} (\cref{tab:supp_baselines}). They showcase that Conf-OT consistently performs better across various scenarios, especially under demanding premises of low error rates ($\alpha=0.05$). Regarding $\text{TIM}_{\text{KL}(\widehat{\kmarg}||\uu_K)}$, we recall that this option provided better set efficiency compared to Conf-OT using APS in the narrower scenario introduced in \cref{tab:supp_baselines}. Nevertheless, these supplementary results show this trend is not generalizable across additional backbones and coverage levels. For example, for ResNet-50 in \cref{tab:supp_baselines}, $\text{TIM}_{\text{KL}(\widehat{\kmarg}||\uu_K)}$ combined with APS produces set sizes nearly $15\%$ larger than Conf-OT, for both $\alpha=0.10$, and $\alpha=0.05$.

\subsection{Further exploration on accuracy against set size improvement}
\label{appendix_results_set_accuracy}

In the main manuscript, we questioned whether the improvements provided by the proposed transductive approach were due solely to better discriminative performance (\cref{main:subsection_ablation}, \cref{fig:accuracy_vs_size}). In this section, we provided additional observations on the discriminative and conformal inference improvements provided by Conf-OT.

\begin{figure}[!ht]
    \begin{center}

        \begin{tabular}{c}
        \hspace{-5mm}\includegraphics[width=.485\linewidth]{figures/corr_size_acc_01_th.pdf} \\
        (a) LAC \cite{lac} \\
        \end{tabular}
        
        \begin{tabular}{cc}
         \hspace{-5mm} \includegraphics[width=.485\linewidth]{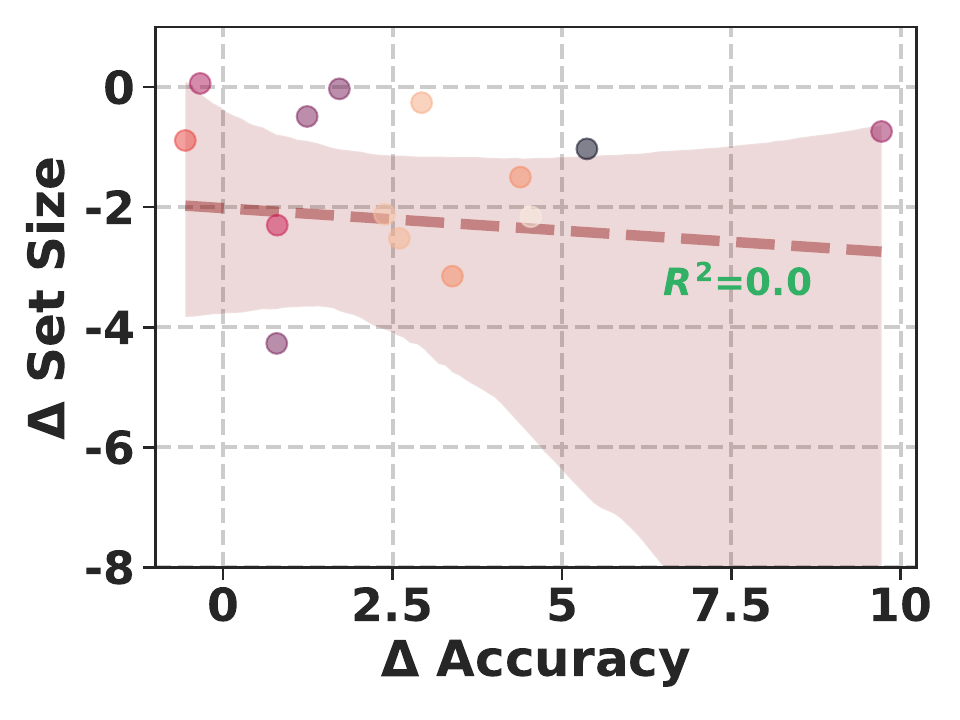} & \includegraphics[width=.485\linewidth]{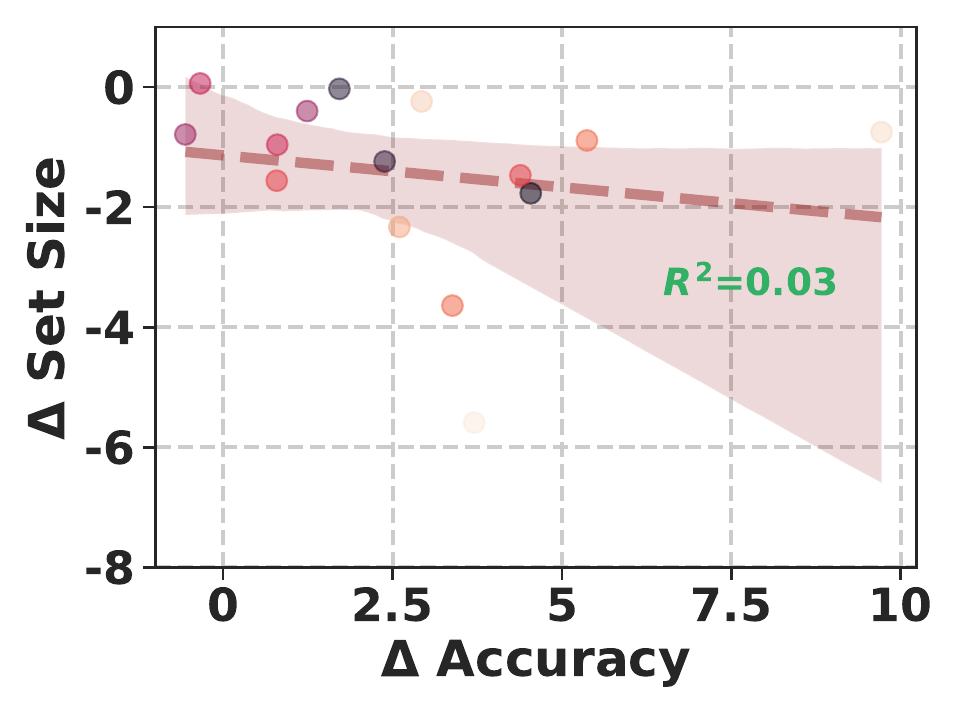}\\
        (b) APS \cite{aps}  & (c) RAPS \cite{raps} \\
        \end{tabular}
        
        \vspace{-2mm}
        \caption{\textbf{Correlation accross datasets of accuracy \textit{vs.} set size change ($\Delta$) increment using Conf-OT}, with popular non-conformal scores, i.e., LAC \cite{lac}, APS \cite{aps}, and RAPS \cite{raps}. Results using CLIP ViT-B/16 on 15 datasets with $\alpha=0.10$. These results complement \cref{fig:accuracy_vs_size} in the main manuscript.}
        \label{fig:accuracy_vs_size_appendix}
    \end{center}
\end{figure}

\mypar{Relative improvements across datasets.} We provide in \cref{fig:accuracy_vs_size_appendix} an extended version of \cref{fig:accuracy_vs_size}(a). These visualizations indicate a limited correlation between the accuracy improvement and set size decreasing for all non-conformity scores employed, both adaptive and non-adaptive measures, i.e., LAC, APS, and RAPS.

\begin{figure}[h!]
    \begin{center}
        \begin{tabular}{cccc}

        \hspace{-5mm}
         \includegraphics[width=.49\linewidth]{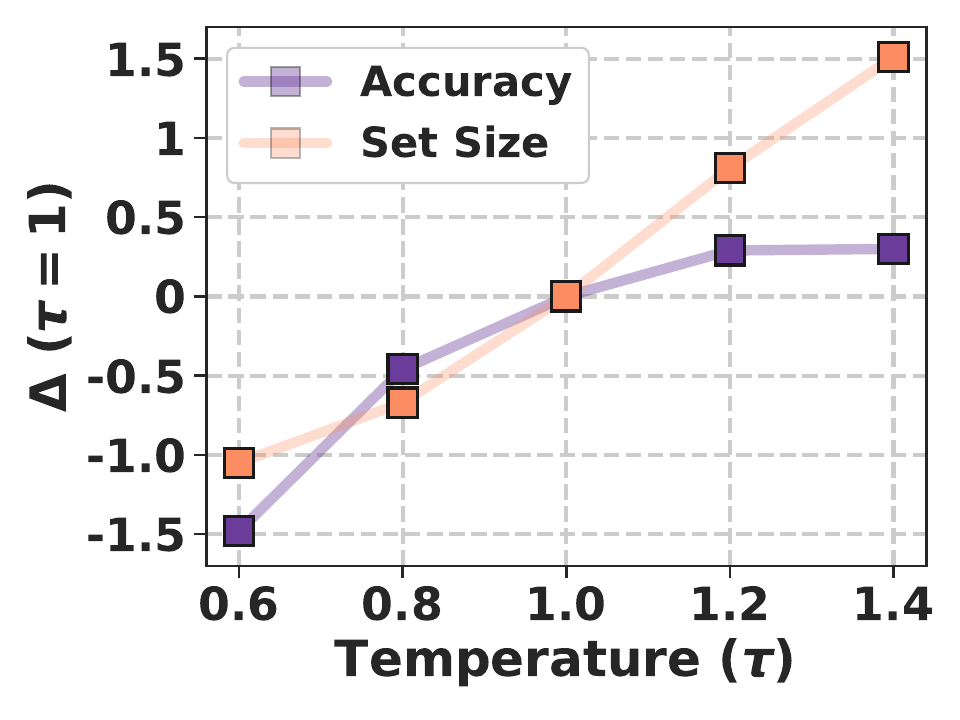} &
         \includegraphics[width=.49\linewidth]{figures/tau_raps_acc_size.pdf} &\\

        (a) APS \cite{aps} & (b) RAPS \cite{raps} & \\

        \end{tabular}
        \vspace{-2mm}
        \caption{\textbf{Relation of accuracy against set size improvement on adaptive scores}, i.e., APS \cite{aps} (a) and RAPS \cite{raps} (b), resulting from modifying the distribution sharpness, via temperature scaling. Results using ViT-B/16 on 15 datasets with $\alpha=0.10$. These results complement \cref{fig:accuracy_vs_size} in the main manuscript.}
        \label{fig:acc_size_tau_supp}
    \end{center}
\end{figure}

\mypar{Effect of temperature scaling in adaptive non-conformity scores.} In the main manuscript (\cref{main:subsection_ablation}), we explored the positive effect of increasing confidence in prediction for adaptive methods using temperature scaling, as observed in \cite{confts}. We now recover this argument to explore how this affects accuracy, showcased in \cref{fig:acc_size_tau_supp}. Such results indicate that, if decreasing the temperature scaling value $\tau$, the efficiency improvement comes at the cost of an accuracy detriment when using Conf-OT. This, again, suggests a disjoint behaviour between discriminative and conformal figures of merit.

\subsection{Conf-OT hyper-parameter studies}
\label{appendix_results_hyperparam}

\begin{figure}[!ht]
    \begin{center}
         \includegraphics[width=.6\linewidth]{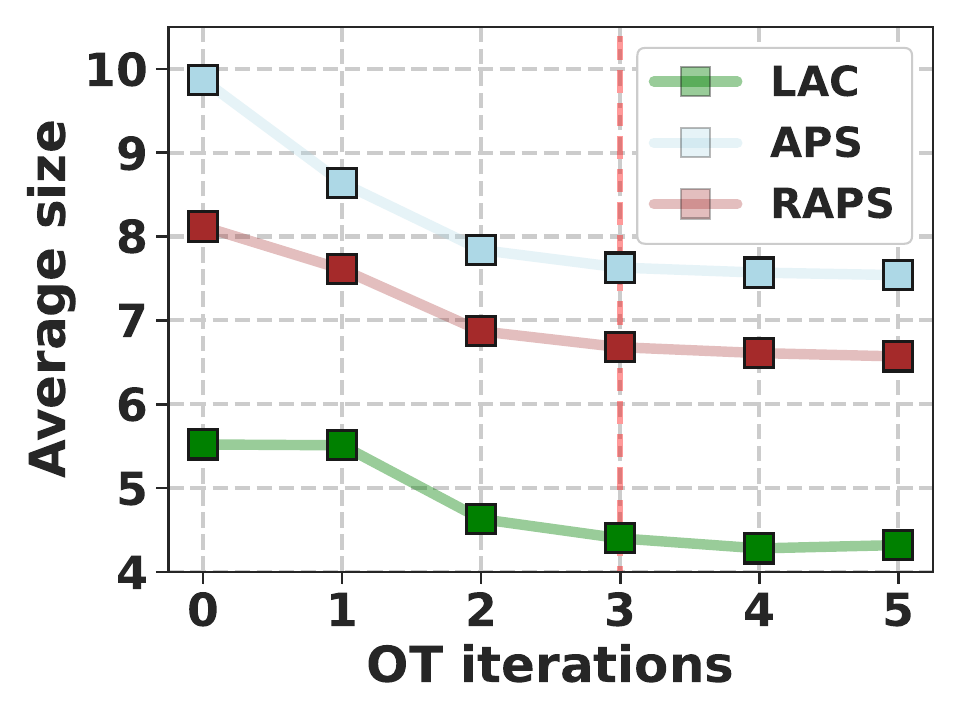} 
        \vspace{-2mm}
        \caption{\textbf{Study on the number of iterations of Conf-OT}. Results using CLIP ViT-B/16 on 15 datasets. Dashed lines indicate the chosen value for this hyper-parameter, \textit{i.e.} $T=3$ repetitions.
        }
        \label{fig:repetitions}
    \end{center}
    \vspace{-5mm}
\end{figure}

\cref{fig:repetitions} presents the convergence of the Sinkhorn algorithm in Conf-OT regarding the number of iterations by measuring set size. These results demonstrate that such algorithms reach a satisfactory convergence after three iterations. This observation is consistent with typical values employed on prior works using Sinkhorn optimal transport in vision tasks \cite{caron2018deep}. Thus, we kept $T=3$ across all experiments and configurations in our experiments.

\subsection{Details on computational efficiency}
\label{appendix_results_comp_efficiency}

The following section includes additional details on the hardware employed for our experiments and the computational efficiency of Conf-OT.

\mypar{Hardware.} All our experiments were conducted on a single GeForce RTX 3060. We extracted the vision features and class text-driven prototypes using GPU resources. However, we evaluated Conf-OT using solely CPU hardware. If available, Conf-OT requires less than 0.7 Gb of memory in a commodity GPU for the most demanding datasets, such as ImageNet. In this scenario, Sinkhorn optimal transport computation speed-up in a factor $\times 10$.

\begin{figure}[!ht]
    \begin{center}
         \includegraphics[width=.95\linewidth]{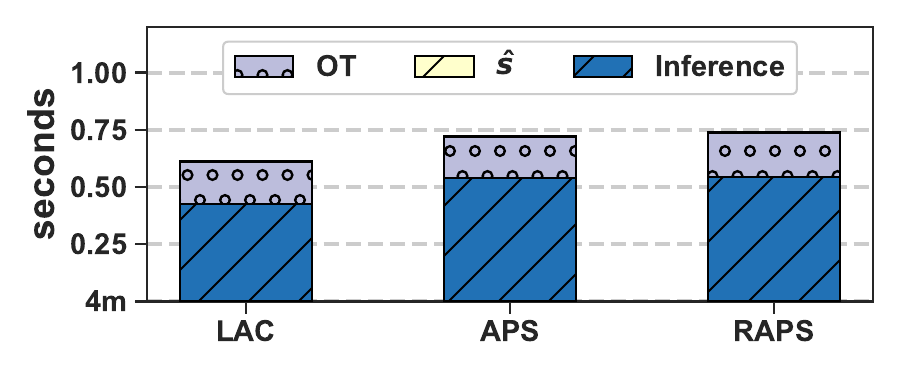} 
        \vspace{-2mm}
        \caption{\textbf{Runtimes of Conf-OT}. The baseline ($4$ minutes) indicates feature extraction runtime. The required OT adaptation times are negligible  ($\leq$ 10 ms) compared to the baseline and inference speed. “$\hat{s}$" indicates finding the conformity score from calibration data. Average results using CLIP ViT-B/16 on 15 datasets.}
        \label{fig:runtimes}
    \end{center}
\end{figure}

\mypar{Runtime analysis in Conf-OT.} \cref{fig:runtimes} provides detailed runtimes for each of the stages in Conf-OT. Note that the baseline time refers to the feature extraction runtime when extracting feature embedding from the vision encoder (nearly $4$ minutes). We recall that Conf-OT is equipped with three stages: \textit{i}) transfer learning trough transductive optimal transport, \textit{ii}) searching the $1-\alpha$ quantile from a non-conformity score distribution in the calibration set, and \textit{iii}) inference on the query samples, which consists of producing the output sets. On average, across 15 tasks, these three stages require less than 0.75 seconds for the full dataset. The results showcase that the latest step involves the main part of the runtime, which is also required if not following our transfer learning pipeline.
On the other hand, threshold computing requires negligible computing times, and the optimal transport adaptation consumes nearly 1/3 of the whole runtime. These results showcase that Conf-OT is extremely efficient, compensating for the necessity of optimizing the optimal transport problem and finding the non-conformity score threshold for each query batch, e.g., if the inference is carried out in small testing batches of $8$ images, the whole Conf-OT procedure should be computed for each batch. Also, it is worth mentioning that these processing times are orders of magnitude smaller than the feature extraction process. Also, \textit{Conf-OT runtimes can be notably decreased using specialized hardware, such as GPUs, for the Sinkhorn algorithm}.

\subsection{Details on data efficiency}
\label{appendix_results_data_efficiency}

The proposed Conf-OT is transductive. That means that access to both calibration and test data during inference is required to perform the transfer learning adaptation between source and target domains. In the following, we explore how sensitive Conf-OT is in two measures: \textit{i}) calibration data requirements, and \textit{ii}) size of the input query batch.

\begin{table}[!ht]
\setlength{\tabcolsep}{3pt}
\centering
\scriptsize
\begin{tabular}{lcccccc}
\toprule
\multicolumn{1}{c}{\multirow{1}{*}{Method}} & \multicolumn{1}{c}{\multirow{1}{*}{Ratio}}  & & \multicolumn{3}{c}{$\alpha$ = 0.10}\\ \cmidrule{4-6}
&  Calib\hspace{0.25mm} -\hspace{0.25mm} Test      & Top-1$\uparrow$       & Cov.   & Size$\downarrow$ & CCV$\downarrow$  \\
\midrule 
\multicolumn{1}{l}{\multirow{4}{*}{LAC}}          & 0.1\hspace{0.25mm} -\hspace{0.25mm} 0.9 & 63.8                       & 0.903      & 7.71                       &  9.65                      \\
                                                  & 0.2\hspace{0.25mm} -\hspace{0.25mm} 0.8 & 63.8                       & 0.899      & 5.56                       &  9.80                      \\
                                                  & 0.5\hspace{0.25mm} -\hspace{0.25mm} 0.5 & 63.8                       & 0.899      & 5.52                       & 10.37                      \\
                                                  & 0.8\hspace{0.25mm} -\hspace{0.25mm} 0.2 & 63.8                       & 0.899      & 5.56                       & 11.70                      \\
\hdashline
\multicolumn{1}{l}{\multirow{4}{*}{Conf-OT+LAC}}  & 0.1\hspace{0.25mm} -\hspace{0.25mm} 0.9 & 66.6                       & 0.901      & 4.53                       &  8.73                      \\
                                                  & 0.2\hspace{0.25mm} -\hspace{0.25mm} 0.8 & 66.7                       & 0.899      & 4.39                       &  8.86                      \\
                                                  & 0.5\hspace{0.25mm} -\hspace{0.25mm} 0.5 & 66.7                       & 0.900      & 4.40                       &  9.48                      \\
                                                  & 0.8\hspace{0.25mm} -\hspace{0.25mm} 0.2 & 66.7                       & 0.899      & 4.41                       & 11.12                      \\
\midrule 
\multicolumn{1}{l}{\multirow{4}{*}{APS}}          & 0.1\hspace{0.25mm} -\hspace{0.25mm} 0.9 & 63.8                       & 0.901      & 9.96                       & 7.55                       \\
                                                  & 0.2\hspace{0.25mm} -\hspace{0.25mm} 0.8 & 63.8                       & 0.900      & 9.94                       & 7.65                       \\
                                                  & 0.5\hspace{0.25mm} -\hspace{0.25mm} 0.5 & 63.8                       & 0.900      & 9.87                       & 8.39                       \\
                                                  & 0.8\hspace{0.25mm} -\hspace{0.25mm} 0.2 & 63.8                       & 0.900      & 9.88                       & 10.32                      \\
\hdashline
\multicolumn{1}{l}{\multirow{4}{*}{Conf-OT+APS}}  & 0.1\hspace{0.25mm} -\hspace{0.25mm} 0.9 & 66.6                       & 0.902      & 7.7                        &  6.44                      \\
                                                  & 0.2\hspace{0.25mm} -\hspace{0.25mm} 0.8 & 66.7                       & 0.901      & 7.64                       &  6.59                      \\
                                                  & 0.5\hspace{0.25mm} -\hspace{0.25mm} 0.5 & 66.7                       & 0.899      & 7.64                       &  7.44                      \\
                                                  & 0.8\hspace{0.25mm} -\hspace{0.25mm} 0.2 & 66.7                       & 0.900      & 7.67                       &  9.62                      \\
\midrule 
\multicolumn{1}{l}{\multirow{4}{*}{RAPS}}         & 0.1\hspace{0.25mm} -\hspace{0.25mm} 0.9 & 63.8                       & 0.900      & 8.12                       &  7.67                      \\
                                                  & 0.2\hspace{0.25mm} -\hspace{0.25mm} 0.8 & 63.8                       & 0.900      & 8.10                       &  7.74                      \\
                                                  & 0.5\hspace{0.25mm} -\hspace{0.25mm} 0.5 & 63.8                       & 0.900      & 8.12                       &  8.50                      \\
                                                  & 0.8\hspace{0.25mm} -\hspace{0.25mm} 0.2 & 63.8                       & 0.900      & 8.10                       & 10.37                      \\
\hdashline
\multicolumn{1}{l}{\multirow{4}{*}{Conf-OT+RAPS}} & 0.1\hspace{0.25mm} -\hspace{0.25mm} 0.9 & 66.6                       & 0.901      & 6.73                       &  6.73                      \\
                                                  & 0.2\hspace{0.25mm} -\hspace{0.25mm} 0.8 & 66.7                       & 0.900      & 6.70                       &  6.64                      \\
                                                  & 0.5\hspace{0.25mm} -\hspace{0.25mm} 0.5 & 66.7                       & 0.900      & 6.68                       &  7.48                      \\
                                                  & 0.8\hspace{0.25mm} -\hspace{0.25mm} 0.2 & 66.7                       & 0.899      & 6.70                       &  9.69                      \\
\bottomrule
\end{tabular}
\caption{\textbf{Robustness to different data ratios}. Results using CLIP ViT-B/16 on 15 datasets averaged across 20 seeds.}
\label{tab:data_ratios}
\end{table}

\begin{table}[!ht]
\centering
\scriptsize
\begin{tabular}{lcccccc}
\toprule
\multicolumn{1}{c}{\multirow{1}{*}{Method}} & \multicolumn{1}{c}{\multirow{1}{*}{M}}  & & \multicolumn{3}{c}{$\alpha$ = 0.10}\\ \cmidrule{4-6}
& & Top-1$\uparrow$       & Cov.   & Size$\downarrow$ & CCV$\downarrow$  \\
\midrule 
LAC                                                         & -     & 63.8 & 0.899 & 5.52 & 10.37 \\
\multicolumn{1}{l}{\multirow{1}{*}{\hspace{2mm}w/ Conf-OT}} & Full  & 66.7 & 0.900 & 4.40 &  9.48 \\\hdashline
\multicolumn{1}{l}{\multirow{1}{*}{\hspace{2mm}w/ Conf-OT}} & 32    & 66.5 & 0.898 & 4.43 &  9.66 \\
\multicolumn{1}{l}{\multirow{1}{*}{\hspace{2mm}w/ Conf-OT}} & 16    & 66.5 & 0.898 & 4.43 &  9.67 \\
\multicolumn{1}{l}{\multirow{1}{*}{\hspace{2mm}w/ Conf-OT}} & 8     & 66.6 & 0.898 & 4.42 &  9.67 \\
\midrule
APS                                                         & -     & 63.8 & 0.900 & 9.87 &  8.39 \\
\multicolumn{1}{l}{\multirow{1}{*}{\hspace{2mm}w/ Conf-OT}} & Full  & 66.7 & 0.899 & 7.64 &  7.44 \\ \hdashline
\multicolumn{1}{l}{\multirow{1}{*}{\hspace{2mm}w/ Conf-OT}} & 32    & 66.5 & 0.900 & 7.68 &  7.51 \\
\multicolumn{1}{l}{\multirow{1}{*}{\hspace{2mm}w/ Conf-OT}} & 16    & 66.5 & 0.900 & 7.68 &  7.52 \\
\multicolumn{1}{l}{\multirow{1}{*}{\hspace{2mm}w/ Conf-OT}} & 8     & 66.6 & 0.900 & 7.67 &  7.54 \\
\midrule
RAPS                                                        & -     & 63.8 & 0.900 & 8.12 &  8.50 \\
\multicolumn{1}{l}{\multirow{1}{*}{\hspace{2mm}w/ Conf-OT}} & Full  & 66.7 & 0.899 & 6.70 &  7.48 \\ \hdashline
\multicolumn{1}{l}{\multirow{1}{*}{\hspace{2mm}w/ Conf-OT}} & 32    & 66.5 & 0.900 & 6.72 &  7.58 \\
\multicolumn{1}{l}{\multirow{1}{*}{\hspace{2mm}w/ Conf-OT}} & 16    & 66.5 & 0.900 & 6.71 &  7.57 \\
\multicolumn{1}{l}{\multirow{1}{*}{\hspace{2mm}w/ Conf-OT}} & 8     & 66.6 & 0.900 & 6.72 &  7.57 \\ 
\bottomrule
\end{tabular}
\caption{\textbf{Robustness to small query batches}. Results using CLIP ViT-B/16 on 15 datasets averaged across 20 seeds. “M" indicates the size of the query batch for Conf-OT. Metrics are extracted by concatenating the predicted sets on the whole test subset.}
\label{tab:small_batch}
\end{table}

\mypar{Robustness to calibration data.} \cref{tab:data_ratios} provides the results of base non-conformal scores over CLIP's zero-shot predictions and atop our Conf-OT method for different calibration/testing data ratios. Results demonstrate a constant performance regarding the efficiency improvements derived from Conf-OT across all these settings and non-conformity scores, even in challenging scenarios such as using solely 10$\%$ of data for calibration.

\mypar{Robustness to small query sets.} \cref{tab:small_batch} contains the results when inference in Conf-OT is performed using extremely small mini-batches of images, e.g., 8, 16, or 32 images, sequentially. The figures of merit show consistent efficiency and class-conditional coverage improvements w.r.t. the base version of each non-conformal score, at the same level as the results observed when using the full batch (all testing samples) simultaneously, as in the main manuscript. As expected, the total runtimes for the whole dataset increase if small mini-batches are used since Conf-OT requires optimization for each batch.

\newpage

\begin{table*}[!ht]
\centering
\scriptsize
\begin{tabular}{clcccccccc}
\toprule
& \multicolumn{1}{c}{\multirow{2}{*}{Method}} & & \multicolumn{3}{c}{$\alpha=0.10$} &  & \multicolumn{3}{c}{$\alpha=0.05$}   \\ \cmidrule{4-6}\cmidrule{8-10}  
&  \multicolumn{1}{c}{}      & Top-1$\uparrow$       & Cov.   & Size$\downarrow$ & CCV$\downarrow$ & & Cov.   & Size$\downarrow$ & CCV$\downarrow$  \\
\midrule 
\multicolumn{1}{c}{\multirow{7}{*}{\rotatebox{90}{
\begin{tabular}[c]{@{}c@{}} \textbf{CLIP} \\ \textbf{ResNet-101} \end{tabular}
}}}
& LAC \cite{lac}                        & 56.7                       & 0.899      & 9.1                         & 10.31                      &      & 0.950      & 16.43                       & 6.14 \\
& \our \hspace{2mm}w/ Conf-OT           & \our \best{59.8}\imp{+3.1} & \our 0.900 & \our \best{7.22}\imp{-1.9}  & \our \best{9.48}\imp{-0.8} & \our & \our 0.950 & \our \best{12.99}\imp{-3.4} & \our \best{5.75}\imp{-0.4} \\
\cmidrule{2-10}  
& APS \cite{aps}                        & 56.7                       & 0.900      & 14.97                       & 8.67                       &      & 0.950      & 24.76                       & 5.55 \\
& \our \hspace{2mm}w/ Conf-OT           & \our \best{59.8}\imp{+3.1} & \our 0.900 & \our \best{11.4}\imp{-3.6}  & \our \best{7.74}\imp{-0.9} & \our & \our 0.950 & \our \best{18.69}\imp{-6.1} & \our \best{5.15}\imp{-0.4} \\
\cmidrule{2-10} 
& RAPS \cite{raps}                      & 56.7                       & 0.901      & 12.12                       & 8.79                       &      & 0.950      & 19.12                       & 5.60 \\
& \our \hspace{2mm}w/ Conf-OT           & \our \best{59.8}\imp{+3.1} & \our 0.900 & \our \best{9.93}\imp{-2.2}  & \our \best{7.82}\imp{-1.0} & \our & \our 0.950 & \our \best{15.27}\imp{-3.9} & \our \best{5.23}\imp{-0.4} \\
\midrule 
\multicolumn{1}{c}{\multirow{7}{*}{\rotatebox{90}{
\begin{tabular}[c]{@{}c@{}} \textbf{CLIP} \\ \textbf{ViT-B/32} \end{tabular}
}}}
& LAC \cite{lac}                        & 58.7                       & 0.900      & 8.20                        & 10.40                      &      & 0.950      & 15.21                       & 6.12 \\
& \our \hspace{2mm}w/ Conf-OT           & \our \best{61.3}\imp{+2.6} & \our 0.899 & \our \best{6.63}\imp{-1.6}  & \our \best{9.39}\imp{-1.0} & \our & \our 0.950 & \our \best{12.05}\imp{-3.2} & \our \best{5.76}\imp{-0.4} \\
\cmidrule{2-10}  
& APS \cite{aps}                        & 58.7                       & 0.901      & 13.55                       & 8.59                       &      & 0.950      & 22.76                       & 5.54 \\
& \our \hspace{2mm}w/ Conf-OT           & \our \best{61.3}\imp{+2.6} & \our 0.900 & \our \best{10.69}\imp{-2.9} & \our \best{7.59}\imp{-1.0} & \our & \our 0.950 & \our \best{17.49}\imp{-5.3} & \our \best{5.05}\imp{-0.5} \\
\cmidrule{2-10} 
& RAPS \cite{raps}                      & 58.7                       & 0.901      & 11.00                       & 8.70                       &      & 0.950      & 18.09                       & 5.61 \\
& \our \hspace{2mm}w/ Conf-OT           & \our \best{61.3}\imp{+2.6} & \our 0.899 & \our \best{9.25}\imp{-1.75}  & \our \best{7.66}\imp{-1.0} & \our & \our 0.950 & \our \best{14.10}\imp{-4.0} & \our \best{5.13}\imp{-0.5} \\
\midrule 
\multicolumn{1}{c}{\multirow{7}{*}{\rotatebox{90}{
\begin{tabular}[c]{@{}c@{}} \textbf{MetaCLIP} \\ \textbf{ViT-B/16} \end{tabular}
}}} 
& LAC \cite{lac}                                    & 69.8                       & 0.900      & 3.79                        & 10.04                       &      & 0.950      & 7.08                        & 5.92 \\
& \our \hspace{2mm}w/ Conf-OT            & \our \best{71.8}\imp{+2.0} & \our 0.900 & \our \best{3.06}\imp{-0.7}  & \our \best{9.47}\imp{-0.6}  & \our & \our 0.950 & \our \best{5.57}\imp{-1.5} & \our \best{5.63}\imp{-0.3} \\
\cmidrule{2-10}  
& APS \cite{aps}                                    & 69.8                       & 0.900      & 7.28                        & 7.76                        &      & 0.949      & 12.4                        & 5.13 \\
& \our \hspace{2mm}w/ Conf-OT            & \our \best{71.8}\imp{+2.0} & \our 0.900 & \our \best{5.58}\imp{-1.7}  & \our \best{7.12}\imp{-0.6}  & \our & \our 0.950 & \our \best{9.80}\imp{-2.6} & \our \best{4.85}\imp{-0.3} \\
\cmidrule{2-10} 
& RAPS \cite{raps}                                   & 69.8                       & 0.900      & 6.03                        & 7.83                        &      & 0.950      & 9.21                        & 5.16 \\
& \our \hspace{2mm}w/ Conf-OT            & \our \best{71.8}\imp{+2.0} & \our 0.900 & \our \best{5.11}\imp{-0.9}  & \our \best{7.13}\imp{-0.7}  & \our & \our 0.950 & \our \best{7.71}\imp{-1.5} & \our \best{4.90}\imp{-0.3} \\
\midrule 
\multicolumn{1}{c}{\multirow{7}{*}{\rotatebox{90}{
\begin{tabular}[c]{@{}c@{}} \textbf{CLIP} \\ \textbf{ViT-L/14} \end{tabular}
}}} 
& LAC \cite{lac}                        & 72.6                       & 0.900      & 2.93                        & 10.34                      &      & 0.950      & 5.36                        & 6.03 \\
& \our \hspace{2mm}w/ Conf-OT           & \our \best{74.7}\imp{+2.1} & \our 0.900 & \our \best{2.5}\imp{-0.4}   & \our \best{9.97}\imp{-0.4} & \our & \our 0.949 & \our \best{4.49}\imp{-0.9}  & \our \best{5.93}\imp{-0.1} \\
\cmidrule{2-10}  
& APS \cite{aps}                        & 72.6                       & 0.900      & 6.40                        & 7.80                       &      & 0.949      & 11.27                       & 5.21 \\
& \our \hspace{2mm}w/ Conf-OT           & \our \best{74.7}\imp{+2.1} & \our 0.900 & \our \best{4.87}\imp{-1.5}  & \our \best{6.98}\imp{-0.8} & \our & \our 0.949 & \our \best{8.39}\imp{-2.9}  & \our \best{4.84}\imp{-0.4} \\
\cmidrule{2-10} 
& RAPS \cite{raps}                      & 72.6                       & 0.900      & 4.93                        & 7.87                       &      & 0.949      & 7.31                        & 5.27 \\
& \our \hspace{2mm}w/ Conf-OT           & \our \best{74.7}\imp{+2.1} & \our 0.900 & \our \best{4.13}\imp{-0.8}  & \our \best{7.01}\imp{-0.9} & \our & \our 0.950 & \our \best{6.22}\imp{-1.1} & \our \best{4.89}\imp{-0.4} \\
\midrule 
\multicolumn{1}{c}{\multirow{7}{*}{\rotatebox{90}{
\begin{tabular}[c]{@{}c@{}} \textbf{MetaCLIP} \\ \textbf{ViT-H/14} \end{tabular}
}}} 
& LAC \cite{lac}                                    & 79.4                       & 0.900      & 1.79                        & 10.91                       &      & 0.950      & 2.98                        & 6.17 \\
& \our \hspace{2mm}w/ Conf-OT            & \our \best{81.1}\imp{+1.7} & \our 0.900 & \our \best{1.59}\imp{-0.2}  & \our \best{10.07}\imp{-0.8} & \our & \our 0.950 & \our \best{2.53}\imp{-0.4} & \our \best{5.92}\imp{-0.3} \\
\cmidrule{2-10}  
& APS \cite{aps}                                    & 79.4                       & 0.900      & 4.45                        & 7.48                        &      & 0.950      & 7.70                        & 4.97 \\
& \our \hspace{2mm}w/ Conf-OT            & \our \best{81.1}\imp{+1.7} & \our 0.899 & \our \best{3.47}\imp{-1.0}  & \our \best{6.60}\imp{-0.9}  & \our & \our 0.949 & \our \best{6.00}\imp{-1.7} & \our \best{4.55}\imp{-0.4} \\
\cmidrule{2-10}
& RAPS \cite{raps}                                   & 79.4                       & 0.900      & 3.45                        & 7.53                        &      & 0.950      & 4.90                        & 5.01 \\
&\our \hspace{2mm}w/ Conf-OT            & \our \best{81.1}\imp{+1.7} & \our 0.899 & \our \best{2.93}\imp{-0.5}  & \our \best{6.61}\imp{-0.9}  & \our & \our 0.950 & \our \best{4.30}\imp{-0.6} & \our \best{4.57}\imp{-0.4} \\
\bottomrule
\end{tabular}
\caption{\textbf{Performance for additional CLIP and MetaCLIP backbones}. Conf-OT performance above popular conformal inference methods such as LAC \cite{lac}, APS \cite{aps}, and RAPS \cite{raps}. Average performance across 15 datasets. We repeat each experiment 20 times. “$\downarrow$" indicates smaller values are better. \textbf{Bold} numbers are superior results. These results complement \cref{main_results} in the main manuscript.}
\label{supp_results_backbones}
\end{table*}

\begin{table*}[!ht]
\centering
\scriptsize
\begin{tabular}{clcccccccc}
\toprule
& \multicolumn{1}{c}{\multirow{2}{*}{Method}} & & \multicolumn{3}{c}{$\alpha=0.10$} &  & \multicolumn{3}{c}{$\alpha=0.05$}   \\ \cmidrule{4-6}\cmidrule{8-10}  
&  \multicolumn{1}{c}{}      & Top-1$\uparrow$       & Cov.   & Size$\downarrow$ & CCV$\downarrow$ & & Cov.   & Size$\downarrow$ & CCV$\downarrow$  \\
\midrule 
\multicolumn{1}{c}{\multirow{15}{*}{\rotatebox{90}{\textbf{CLIP ResNet-50}}}}
& LAC \cite{lac}                                             & 54.7                       & 0.900          & 10.77                       & 9.82                       &      & 0.950      & 19.22                       & 5.91 \\
& \hspace{2mm}$\text{TIM}_{\text{KL}(\widehat{\kmarg}||\uu_K)}$ \cite{tim}   & 55.7\imp{+1.0}             & 0.899          & 14.45\wor{+3.7}             & 9.69\imp{-0.1}             &      & 0.950      & 26.31\wor{+7.1}             & 5.70\imp{-0.2} \\
& \hspace{2mm}$\text{TIM}_{\text{KL}(\widehat{\kmarg}||\kmarg)}$ \cite{tim}   & 56.4\imp{+1.7}             & 0.899          & 13.63\wor{+2.9}             & 12.02\wor{+2.2}            &      & 0.950      & 24.88\wor{+5.7}             & 6.03\wor{+0.1} \\
& \hspace{2mm}TransCLIP \cite{zanella2024boosting}                           & 55.7\imp{+1.0}             & \nc{0.861}     & \best{7.83}\imp{-2.9}       & 12.20\wor{+2.4}            &      & \nc{0.881} & \best{8.81}\imp{-10.4}      & 11.03\wor{+5.1} \\
& \our \hspace{2mm}Conf-OT                        & \our \best{57.3}\imp{+2.6} & \our 0.900     & \our 8.61\imp{-2.1}  & \our \best{9.15}\imp{-0.6} & \our & \our 0.951 & \our 15.53\imp{-3.6} & \our \best{5.61}\imp{-0.3} \\
\cmidrule{2-10}  
& APS \cite{aps}                                             & 54.7                       & 0.900          & 16.35                       & 8.36                       &      & 0.950      & 26.50                       & 5.34 \\
& \hspace{2mm}$\text{TIM}_{\text{KL}(\widehat{\kmarg}||\uu_K)}$ \cite{tim}   & 55.7\imp{+1.0}             & 0.900          & 13.33\imp{-3.0}             & 8.64\wor{+0.3}             &      & 0.950      & 24.73\imp{-1.8}             & 5.37\wor{+0.0} \\
& \hspace{2mm}$\text{TIM}_{\text{KL}(\widehat{\kmarg}||\kmarg)}$ \cite{tim}   & 56.4\imp{+1.7}             & 0.900          & 13.26\imp{-3.1}             & 8.69\wor{+0.3}             &      & 0.950      & 24.09\imp{-2.4}             & 5.47\wor{+0.1} \\
& \hspace{2mm}TransCLIP \cite{zanella2024boosting}                           & 55.7\imp{+1.0}             & \nc{0.873}     & 39.05\wor{+22.7}            & 11.15\wor{+2.8}            &      & \nc{0.899} & 46.78\wor{+20.3}            & 9.41\wor{+4.1} \\
& \our \hspace{2mm}Conf-OT                        & \our \best{57.3}\imp{+2.6} & \our 0.900     & \our \best{12.94}\imp{-3.4} & \our \best{7.64}\imp{-0.7} & \our & \our 0.950 & \our \best{20.96}\imp{-5.5} & \our \best{5.03}\imp{-0.3} \\
\cmidrule{2-10} 
& RAPS \cite{raps}                                            & 54.7                       & 0.900          & 13.37                       & 8.46                       &      & 0.950      & 22.06                       & 5.44 \\
& \hspace{2mm}$\text{TIM}_{\text{KL}(\widehat{\kmarg}||\uu_K)}$ \cite{tim}   & 55.7\imp{+1.0}             & 0.900          & 13.18\imp{-0.2}             & 8.64\wor{+0.2}             &      & 0.950      & 24.54\wor{+2.5}             & 5.38\imp{-0.1} \\
& \hspace{2mm}$\text{TIM}_{\text{KL}(\widehat{\kmarg}||\kmarg)}$ \cite{tim}   & 56.4\imp{+1.7}             & 0.900          & 12.99\imp{-0.4}             & 8.71\wor{+0.3}             &      & 0.950      & 23.59\wor{+1.5}             & 5.51\wor{+0.1} \\
& \hspace{2mm}TransCLIP \cite{zanella2024boosting}                           & 55.7\imp{+1.0}             & 0.900          & 13.68\wor{+0.3}             & 9.94\wor{+1.5}             &      & 0.949      & 28.16\wor{+6.1}             & 6.03\wor{+0.6} \\
& \our \hspace{2mm}Conf-OT                        & \our \best{57.3}\imp{+2.6} & \our 0.900 & \our \best{11.17}\imp{-2.2} & \our \best{7.72}\imp{-0.7} & \our & \our 0.950 & \our \best{17.24}\imp{-4.8} & \our \best{5.19}\imp{-0.2} \\
\midrule 
\multicolumn{1}{c}{\multirow{15}{*}{\rotatebox{90}{\textbf{CLIP ViT-B/16}}}} 
& LAC \cite{lac}                                             & 63.8                       & 0.899          & 5.52                        & 10.37                       &      & 0.950      & 10.24                       & 6.14 \\
& \hspace{2mm}$\text{TIM}_{\text{KL}(\widehat{\kmarg}||\uu_K)}$ \cite{tim}   & 64.7\imp{+0.9}             & 0.899          & 8.30\wor{+2.8}              & 10.41\wor{+0.0}             &      & 0.950      & 15.89\wor{+5.7}             & 6.03\imp{-0.1} \\
& \hspace{2mm}$\text{TIM}_{\text{KL}(\widehat{\kmarg}||\kmarg)}$ \cite{tim}   & 65.0\imp{+1.2}             & 0.898          & 7.73\wor{+2.2}              & 10.89\wor{+0.5}             &      & 0.950      & 14.68\wor{+4.4}             & 6.40\wor{+0.3} \\
& \hspace{2mm}TransCLIP \cite{zanella2024boosting}                           & 65.1\imp{+1.3}             & \nc{0.892}     & 5.76\wor{+0.2}              & 11.02\wor{+0.7}             &      & \nc{0.921} & \best{7.02}\imp{-3.2}       & 8.31\wor{+2.2} \\
& \our \hspace{2mm}Conf-OT                        & \our \best{66.7}\imp{+2.9} & \our 0.900 & \our \best{4.40}\imp{-1.1} & \our \best{9.48}\imp{-0.8} & \our & \our 0.949 & \our 7.99\imp{-2.2} & \our \best{5.80}\imp{-0.3} \\
\cmidrule{2-10}  
& APS \cite{aps}                                             & 63.8                       & 0.900          & 9.87                        & 8.39                        &      & 0.950      & 16.92                       & 5.51 \\
& \hspace{2mm}$\text{TIM}_{\text{KL}(\widehat{\kmarg}||\uu_K)}$ \cite{tim}   & 64.7\imp{+0.9}             & 0.900          & \best{7.24}\imp{-2.6}       & 9.32\wor{+0.9}              &      & 0.950      & 14.03\imp{-2.9}             & 5.88\wor{+0.4} \\
& \hspace{2mm}$\text{TIM}_{\text{KL}(\widehat{\kmarg}||\kmarg)}$ \cite{tim}   & 65.0\imp{+1.2}             & 0.900          & 7.82\imp{-2.1}              & 9.38\wor{+1.0}              &      & 0.950      & 14.34\imp{-2.6}             & 6.03\wor{+0.5} \\
& \hspace{2mm}TransCLIP \cite{zanella2024boosting}                           & 65.1\imp{+1.3}             & \nc{0.892}     & 8.27\imp{-1.6}              & 11.50\wor{+3.1}             &      & \nc{0.931} & 38.86\wor{+21.9}            & 7.47\wor{+2.0} \\
& \our \hspace{2mm}Conf-OT                        & \our \best{66.7}\imp{+2.9} & \our 0.899 & \our 7.64\imp{-2.2} & \our \best{7.44}\imp{-0.9} & \our & \our 0.949 & \our \best{12.58}\imp{-4.3} & \our \best{5.09}\imp{-0.4} \\
\cmidrule{2-10} 
& RAPS \cite{raps}                                            & 63.8                       & 0.900          & 8.12                        & 8.50                        &      & 0.950      & 12.66                       & 5.52 \\
& \hspace{2mm}$\text{TIM}_{\text{KL}(\widehat{\kmarg}||\uu_K)}$ \cite{tim}   & 64.7\imp{+0.9}             & 0.900          & 7.18\imp{-0.9}              & 9.32\wor{+0.8}              &      & 0.950      & 13.92\wor{+1.3}             & 5.99\wor{+0.5} \\
& \hspace{2mm}$\text{TIM}_{\text{KL}(\widehat{\kmarg}||\kmarg)}$ \cite{tim}   & 65.0\imp{+1.2}             & 0.900          & 7.68\imp{-0.4}              & 9.42\wor{+0.9}              &      & 0.950      & 14.11\wor{+1.5}             & 6.04\wor{+0.5} \\
& \hspace{2mm}TransCLIP \cite{zanella2024boosting}                           & 65.1\imp{+1.3}             & 0.899          & 7.17\imp{-1.0}              & 10.20\wor{+1.7}             &      & 0.949      & 15.26\wor{+2.6}             & 6.31\wor{+0.8} \\
& \our \hspace{2mm}Conf-OT                        & \our \best{66.7}\imp{+2.9} & \our 0.900 & \our \best{6.68}\imp{-1.4} & \our \best{7.48}\imp{-1.0} & \our & \our 0.949 & \our \best{10.11}\imp{-2.5} & \our \best{5.16}\imp{-0.3} \\
\bottomrule
\end{tabular}
\caption{\textbf{Comparison of transductive baselines for additional CLIP backbones}. Average performance across 15 datasets. We repeat each experiment 20 times. “$\downarrow$" indicates smaller values are better. \textbf{Bold} numbers are superior results. These results complement \cref{tab:transductive_baselines} in the main manuscript.\nc{Grayish} marginal coverage (“Cov.") indicates an unsatisfied error rate, usually seen in TransCLIP.}
\label{tab:supp_baselines}
\end{table*}

\begin{table}[!ht]
\setlength{\tabcolsep}{3.0pt}
\centering
\scriptsize
\begin{tabular}{lcccccccc}
\toprule
\multicolumn{1}{c}{\multirow{2}{*}{Method}} & & \multicolumn{3}{c}{$\alpha=0.10$} &  & \multicolumn{3}{c}{$\alpha=0.05$}   \\ \cmidrule{3-5}\cmidrule{7-9}  
\multicolumn{1}{c}{}      & Top-1$\uparrow$       & Cov   & Size$\downarrow$ & CCV$\downarrow$ & & Cov.   & Size$\downarrow$ & CCV$\downarrow$  \\
\midrule 
\multicolumn{9}{l}{\textbf{SUN397}} \\ \hdashline
LAC \cite{lac}                                   & 58.7 & 0.899 & 4.28  & 8.90  &  & 0.950 & 8.42  & 5.44 \\
\our \hspace{2mm}w/ Conf-OT                      & \our 62.0 & \our 0.899 & \our 3.62  & \our 7.64  & \our  & \our 0.951 & \our 6.77  & \our 4.83 \\
APS \cite{aps}                                   & 58.7 & 0.901 & 8.98  & 7.59  &  & 0.949 & 16.16 & 5.01 \\
\our \hspace{2mm}w/ Conf-OT                      & \our 62.0 & \our 0.900 & \our 7.05  & \our 6.60  & \our  & \our 0.950 & \our 12.20 & \our 4.46 \\
RAPS \cite{raps}                                 & 58.7 & 0.902 & 8.16  & 7.59  &  & 0.949 & 13.63 & 5.06 \\
\bottomrule
\end{tabular}
\vspace{-2mm}
\caption{Results on fine-grained tasks with CLIP ResNet-50.}
\label{appendix_results_rn50_10tasks}
\end{table}

\begin{table}[!ht]
\setlength{\tabcolsep}{3.0pt}
\centering
\scriptsize
\begin{tabular}{lcccccccc}
\toprule
\multicolumn{1}{c}{\multirow{2}{*}{Method}} & & \multicolumn{3}{c}{$\alpha=0.10$} &  & \multicolumn{3}{c}{$\alpha=0.05$}   \\ \cmidrule{3-5}\cmidrule{7-9}  
\multicolumn{1}{c}{}      & Top-1$\uparrow$       & Cov   & Size$\downarrow$ & CCV$\downarrow$ & & Cov.   & Size$\downarrow$ & CCV$\downarrow$  \\
\midrule 
\multicolumn{9}{l}{\textbf{ImageNet}} \\ \hdashline
LAC \cite{lac}                                   & 60.3 & 0.900 & 5.18  & 8.25  &  & 0.950 & 10.74 & 5.03 \\
\our \hspace{2mm}w/ Conf-OT                      & \our 61.6 & \our 0.901 & \our 4.81  & \our 7.63  & \our  & \our 0.950 & \our 10.18 & \our 4.78 \\ 
APS \cite{aps}                                   & 60.3 & 0.900 & 15.06 & 6.82  &  & 0.950 & 29.62 & 4.66 \\
\our \hspace{2mm}w/ Conf-OT                      & \our 61.6 & \our 0.900 & \our 12.38 & \our 6.32  & \our  & \our 0.950 & \our 23.58 & \our 4.30 \\
RAPS \cite{raps}                                 & 60.3 & 0.899 & 11.23 & 6.89  &  & 0.950 & 18.24 & 4.71 \\
\bottomrule
\end{tabular}
\vspace{-2mm}
\caption{Results on ImageNet shifts using CLIP ResNet-50.}
\label{appendix_results_rn50_imagenet}
\end{table}

\begin{table}[!ht]
\setlength{\tabcolsep}{3.0pt}
\centering
\scriptsize
\begin{tabular}{lcccccccc}
\toprule
\multicolumn{1}{c}{\multirow{2}{*}{Method}} & & \multicolumn{3}{c}{$\alpha=0.10$} &  & \multicolumn{3}{c}{$\alpha=0.05$}   \\ \cmidrule{3-5}\cmidrule{7-9}  
\multicolumn{1}{c}{}      & Top-1$\uparrow$       & Cov   & Size$\downarrow$ & CCV$\downarrow$ & & Cov.   & Size$\downarrow$ & CCV$\downarrow$  \\
\midrule 
\multicolumn{9}{l}{\textbf{SUN397}} \\ \hdashline
LAC \cite{lac}                                   & 58.8 & 0.901 & 4.43  & 9.22  &  & 0.951 & 8.80  & 5.57 \\
\our \hspace{2mm}w/ Conf-OT                      & \our 62.2 & \our 0.900 & \our 3.46  & \our 7.55  & \our  & \our 0.951 & \our 6.67  & \our 4.95 \\
APS \cite{aps}                                   & 58.8 & 0.901 & 9.93  & 8.04  &  & 0.950 & 18.80 & 5.14 \\
\our \hspace{2mm}w/ Conf-OT                      & \our 62.2 & \our 0.900 & \our 7.04  & \our 6.56  & \our  & \our 0.950 & \our 12.39 & \our 4.44 \\
RAPS \cite{raps}                                 & 58.8 & 0.901 & 8.88  & 8.09  &  & 0.950 & 15.01 & 5.22 \\
\bottomrule
\end{tabular}
\vspace{-2mm}
\caption{Results on fine-grained tasks with CLIP ResNet-101.}
\label{appendix_results_rn101_10tasks}
\end{table}

\begin{table}[!ht]
\setlength{\tabcolsep}{3.0pt}
\centering
\scriptsize
\begin{tabular}{lcccccccc}
\toprule
\multicolumn{1}{c}{\multirow{2}{*}{Method}} & & \multicolumn{3}{c}{$\alpha=0.10$} &  & \multicolumn{3}{c}{$\alpha=0.05$}   \\ \cmidrule{3-5}\cmidrule{7-9}  
\multicolumn{1}{c}{}      & Top-1$\uparrow$       & Cov   & Size$\downarrow$ & CCV$\downarrow$ & & Cov.   & Size$\downarrow$ & CCV$\downarrow$  \\
\midrule 
\multicolumn{9}{l}{\textbf{ImageNet}} \\ \hdashline
LAC \cite{lac}                                   & 62.6 & 0.900 & 4.32  & 8.56  &  & 0.950 & 8.98  & 5.13 \\
\our \hspace{2mm}w/ Conf-OT                      & \our 64.1 & \our 0.900 & \our 3.98  & \our 7.85  & \our  & \our 0.950 & \our 8.28  & \our 4.72 \\
APS \cite{aps}                                   & 62.6 & 0.900 & 14.97 & 6.74  &  & 0.950 & 29.28 & 4.51 \\
\our \hspace{2mm}w/ Conf-OT                      & \our 64.1 & \our 0.901 & \our 11.61 & \our 6.35  & \our  & \our 0.950 & \our 22.40 & \our 4.29 \\
RAPS \cite{raps}                                 & 62.6 & 0.901 & 10.87 & 6.84  &  & 0.950 & 17.41 & 4.63 \\
\bottomrule
\end{tabular}
\vspace{-2mm}
\caption{Results on ImageNet shifts with CLIP ResNet-101.}
\label{appendix_results_rn101_imagenet}
\end{table}

\begin{table}[!ht]
\setlength{\tabcolsep}{3.0pt}
\centering
\scriptsize
\begin{tabular}{lcccccccc}
\toprule
\multicolumn{1}{c}{\multirow{2}{*}{Method}} & & \multicolumn{3}{c}{$\alpha=0.10$} &  & \multicolumn{3}{c}{$\alpha=0.05$}   \\ \cmidrule{3-5}\cmidrule{7-9}  
\multicolumn{1}{c}{}      & Top-1$\uparrow$       & Cov   & Size$\downarrow$ & CCV$\downarrow$ & & Cov.   & Size$\downarrow$ & CCV$\downarrow$  \\
\midrule 
\multicolumn{9}{l}{\textbf{SUN397}} \\ \hdashline
LAC \cite{lac}                                   & 61.9 & 0.900 & 3.64  & 9.10   &  & 0.949 & 6.55  & 5.54 \\
\our \hspace{2mm}w/ Conf-OT                      & \our 64.8 & \our 0.898 & \our 3.02  & \our 7.66   & \our  & \our 0.951 & \our 5.36  & \our 4.77 \\ 
APS \cite{aps}                                   & 61.9 & 0.900 & 7.62  & 7.30   &  & 0.950 & 13.52 & 4.85 \\
\our \hspace{2mm}w/ Conf-OT                      & \our 64.8 & \our 0.901 & \our 5.92  & \our 6.27   & \our  & \our 0.949 & \our 9.75  & \our 4.28 \\
RAPS \cite{raps}                                 & 61.9 & 0.900 & 6.85  & 7.36   &  & 0.949 & 11.12 & 4.87 \\
\bottomrule
\end{tabular}
\vspace{-2mm}
\caption{Results on fine-grained tasks with CLIP ViT-B/32.}
\label{appendix_results_vit32_10tasks}
\end{table}

\begin{table}[!ht]
\setlength{\tabcolsep}{3.0pt}
\centering
\scriptsize
\begin{tabular}{lcccccccc}
\toprule
\multicolumn{1}{c}{\multirow{2}{*}{Method}} & & \multicolumn{3}{c}{$\alpha=0.10$} &  & \multicolumn{3}{c}{$\alpha=0.05$}   \\ \cmidrule{3-5}\cmidrule{7-9}  
\multicolumn{1}{c}{}      & Top-1$\uparrow$       & Cov   & Size$\downarrow$ & CCV$\downarrow$ & & Cov.   & Size$\downarrow$ & CCV$\downarrow$  \\
\midrule 
\multicolumn{9}{l}{\textbf{ImageNet}} \\ \hdashline
LAC \cite{lac}                                   & 63.8 & 0.901 & 4.13  & 8.40  &  & 0.950 & 8.61  & 5.00 \\
\our \hspace{2mm}w/ Conf-OT                      & \our 64.7 & \our 0.901 & \our 3.84  & \our 7.57  & \our  & \our 0.950 & \our 8.18  & \our 4.71 \\
APS \cite{aps}                                   & 63.8 & 0.900 & 13.94 & 6.89  &  & 0.950 & 27.61 & 4.52 \\
\our \hspace{2mm}w/ Conf-OT                      & \our 64.7 & \our 0.900 & \our 10.90 & \our 6.14  & \our  & \our 0.950 & \our 20.93 & \our 4.26 \\
RAPS \cite{raps}                                 & 63.8 & 0.901 & 10.15 & 7.00  &  & 0.950 & 16.24 & 4.56 \\
\bottomrule
\end{tabular}
\vspace{-2mm}
\caption{Results on ImageNet shifts with CLIP ViT-B/32.}
\label{appendix_results_vitb32_imagenet}
\end{table}

\begin{table}[!ht]
\setlength{\tabcolsep}{3.0pt}
\centering
\scriptsize
\begin{tabular}{lcccccccc}
\toprule
\multicolumn{1}{c}{\multirow{2}{*}{Method}} & & \multicolumn{3}{c}{$\alpha=0.10$} &  & \multicolumn{3}{c}{$\alpha=0.05$}   \\ \cmidrule{3-5}\cmidrule{7-9}  
\multicolumn{1}{c}{}      & Top-1$\uparrow$       & Cov   & Size$\downarrow$ & CCV$\downarrow$ & & Cov.   & Size$\downarrow$ & CCV$\downarrow$  \\
\midrule 
\multicolumn{9}{l}{\textbf{SUN397}} \\ \hdashline
LAC \cite{lac}                                   & 62.5 & 0.900 & 3.47  & 9.60  &  & 0.951 & 6.52  & 5.68  \\
\our \hspace{2mm}w/ Conf-OT                      & \our 67.0 & \our 0.899 & \our 2.72  & \our 7.81  & \our  & \our 0.951 & \our 4.83  & \our 4.86 \\
APS \cite{aps}                                   & 62.5 & 0.901 & 7.42  & 7.74  &  & 0.951 & 13.70 & 5.03 \\
\our \hspace{2mm}w/ Conf-OT                      & \our 67.0 & \our 0.900 & \our 5.26  & \our 6.42  & \our  & \our 0.951 & \our 9.05  & \our 4.33 \\
RAPS \cite{raps}                                 & 62.5 & 0.902 & 6.70  & 7.90  &  & 0.951 & 10.89 & 5.14 \\
\bottomrule
\end{tabular}
\vspace{-2mm}
\caption{Results on fine-grained tasks with CLIP ViT-B/16.}
\label{appendix_results_vit16_10tasks}
\end{table}

\begin{table}[!ht]
\setlength{\tabcolsep}{3.0pt}
\centering
\scriptsize
\begin{tabular}{lcccccccc}
\toprule
\multicolumn{1}{c}{\multirow{2}{*}{Method}} & & \multicolumn{3}{c}{$\alpha=0.10$} &  & \multicolumn{3}{c}{$\alpha=0.05$}   \\ \cmidrule{3-5}\cmidrule{7-9}  
\multicolumn{1}{c}{}      & Top-1$\uparrow$       & Cov   & Size$\downarrow$ & CCV$\downarrow$ & & Cov.   & Size$\downarrow$ & CCV$\downarrow$  \\
\midrule 
\multicolumn{9}{l}{\textbf{ImageNet}} \\ \hdashline
LAC \cite{lac}                                   & 68.7 & 0.901 & 2.81  & 8.56  &  & 0.950 & 5.51  & 5.20 \\
\our \hspace{2mm}w/ Conf-OT                      & \our 69.5 & \our 0.901 & \our 2.70  & \our 7.71  & \our  & \our 0.950 & \our 5.22  & \our 4.77 \\
APS \cite{aps}                                   & 68.7 & 0.901 & 10.07 & 6.84  &  & 0.950 & 20.65 & 4.54 \\
\our \hspace{2mm}w/ Conf-OT                      & \our 69.5 & \our 0.901 & \our 7.77  & \our 6.32  & \our  & \our 0.950 & \our 15.15 & \our 4.32 \\
RAPS \cite{raps}                                 & 68.7 & 0.901 & 7.34  & 6.92  &  & 0.950 & 11.82 & 4.65 \\
\bottomrule
\end{tabular}
\vspace{-2mm}
\caption{Results on ImageNet shifts with CLIP ViT-B/16.}
\label{appendix_results_vitb16_imagenet}
\end{table}

\begin{table}[!ht]
\setlength{\tabcolsep}{3.0pt}
\centering
\scriptsize
\begin{tabular}{lcccccccc}
\toprule
\multicolumn{1}{c}{\multirow{2}{*}{Method}} & & \multicolumn{3}{c}{$\alpha=0.10$} &  & \multicolumn{3}{c}{$\alpha=0.05$}   \\ \cmidrule{3-5}\cmidrule{7-9}  
\multicolumn{1}{c}{}      & Top-1$\uparrow$       & Cov   & Size$\downarrow$ & CCV$\downarrow$ & & Cov.   & Size$\downarrow$ & CCV$\downarrow$  \\
\midrule 
\multicolumn{9}{l}{\textbf{SUN397}} \\ \hdashline
LAC \cite{lac}                                   & 67.6 & 0.900 & 2.85  & 9.78  &  & 0.950 & 5.2   & 5.51 \\
\our \hspace{2mm}w/ Conf-OT                      & \our 71.0 & \our 0.901 & \our 2.24  & \our 7.70  & \our  & \our 0.950 & \our 3.87  & \our 4.69 \\
APS \cite{aps}                                   & 67.6 & 0.900 & 6.99  & 7.70  &  & 0.951 & 14.12 & 4.95 \\
\our \hspace{2mm}w/ Conf-OT                      & \our 71.0 & \our 0.901 & \our 4.46  & \our 6.11  & \our  & \our 0.951 & \our 8.14  & \our 4.20 \\
RAPS \cite{raps}                                 & 67.6 & 0.901 & 5.96  & 7.77  &  & 0.951 & 10.12 & 5.05 \\
\bottomrule
\end{tabular}
\vspace{-2mm}
\caption{Results on fine-grained tasks with CLIP ViT-L/14.}
\label{appendix_results_vitl14_10tasks}
\end{table}

\begin{table}[!ht]
\setlength{\tabcolsep}{3.0pt}
\centering
\scriptsize
\begin{tabular}{lcccccccc}
\toprule
\multicolumn{1}{c}{\multirow{2}{*}{Method}} & & \multicolumn{3}{c}{$\alpha=0.10$} &  & \multicolumn{3}{c}{$\alpha=0.05$}   \\ \cmidrule{3-5}\cmidrule{7-9}  
\multicolumn{1}{c}{}      & Top-1$\uparrow$       & Cov   & Size$\downarrow$ & CCV$\downarrow$ & & Cov.   & Size$\downarrow$ & CCV$\downarrow$  \\
\midrule 
\multicolumn{9}{l}{\textbf{ImageNet}} \\ \hdashline
LAC \cite{lac}                                   & 75.8 & 0.900 & 1.86 & 8.70  &  & 0.950 & 3.39  & 5.26 \\
\our \hspace{2mm}w/ Conf-OT                      & \our 75.7 & \our 0.901 & \our 1.88 & \our 7.78  & \our  & \our 0.950 & \our 3.48  & \our 4.81 \\ 
APS \cite{aps}                                   & 75.8 & 0.901 & 6.96 & 6.82  &  & 0.949 & 15.22 & 4.63 \\
\our \hspace{2mm}w/ Conf-OT                      & \our 75.7 & \our 0.900 & \our 5.53 & \our 6.25  & \our  & \our 0.949 & \our 11.63 & \our 4.41 \\
RAPS \cite{raps}                                 & 75.8 & 0.900 & 4.86 & 6.86  &  & 0.949 & 7.75  & 4.72 \\
\bottomrule
\end{tabular}
\vspace{-2mm}
\caption{Results on ImageNet shifts with CLIP ViT-L/14.}
\label{appendix_results_vitbl14_imagenet}
\end{table}

\begin{table}[!ht]
\setlength{\tabcolsep}{3.0pt}
\centering
\scriptsize
\begin{tabular}{lcccccccc}
\toprule
\multicolumn{1}{c}{\multirow{2}{*}{Method}} & & \multicolumn{3}{c}{$\alpha=0.10$} &  & \multicolumn{3}{c}{$\alpha=0.05$}   \\ \cmidrule{3-5}\cmidrule{7-9}  
\multicolumn{1}{c}{}      & Top-1$\uparrow$       & Cov   & Size$\downarrow$ & CCV$\downarrow$ & & Cov.   & Size$\downarrow$ & CCV$\downarrow$  \\
\midrule 
\multicolumn{9}{l}{\textbf{SUN397}} \\ \hdashline
LAC \cite{lac}                                   & 69.9 & 0.899 & 2.26  & 8.48  &  & 0.950 & 3.82  & 5.10 \\
\our \hspace{2mm}w/ Conf-OT                      & \our71.7 & \our0.900 & \our2.11  & \our7.40  & \our & \our0.950 & \our3.53  & \our4.62 \\
APS \cite{aps}                                   & 69.9 & 0.899 & 4.23  & 6.66  &  & 0.949 & 7.19  & 4.51 \\
\our \hspace{2mm}w/ Conf-OT                      & \our71.7 & \our0.899 & \our3.70  & \our6.01  & \our & \our0.949 & \our6.05  & \our4.18 \\
RAPS \cite{raps}                                 & 69.9 & 0.899 & 3.97  & 6.74  &  & 0.949 & 6.26  & 4.52 \\
\bottomrule
\end{tabular}
\vspace{-2mm}
\caption{Results on fine-grained tasks with MetaCLIP ViT-B/16.}
\label{appendix_results_MetaCLIP_vitB16_10tasks}
\end{table}

\begin{table}[!ht]
\setlength{\tabcolsep}{3.0pt}
\centering
\scriptsize
\begin{tabular}{lcccccccc}
\toprule
\multicolumn{1}{c}{\multirow{2}{*}{Method}} & & \multicolumn{3}{c}{$\alpha=0.10$} &  & \multicolumn{3}{c}{$\alpha=0.05$}   \\ \cmidrule{3-5}\cmidrule{7-9}  
\multicolumn{1}{c}{}      & Top-1$\uparrow$       & Cov   & Size$\downarrow$ & CCV$\downarrow$ & & Cov.   & Size$\downarrow$ & CCV$\downarrow$  \\
\midrule 
\multicolumn{9}{l}{\textbf{ImageNet}} \\ \hdashline
LAC \cite{lac}                                   & 72.2 & 0.901 & 2.30  & 8.24  &  & 0.950 & 4.45  & 5.19 \\
\our \hspace{2mm}w/ Conf-OT                      & \our72.7 & \our0.901 & \our2.24  & \our7.58  & \our & \our0.951 & \our4.36  & \our4.77 \\
APS \cite{aps}                                   & 72.2 & 0.899 & 8.26  & 6.52  &  & 0.950 & 16.70 & 4.46 \\
\our \hspace{2mm}w/ Conf-OT                      & \our72.7 & \our0.900 & \our6.80  & \our6.16  & \our & \our0.950 & \our13.61 & \our4.23 \\
RAPS \cite{raps}                                 & 72.2 & 0.899 & 6.20  & 6.62  &  & 0.950 & 9.93  & 4.52 \\
\bottomrule
\end{tabular}
\vspace{-2mm}
\caption{Results on ImageNet shifts with MetaCLIP ViT-B/16.}
\label{appendix_results_MetaCLIP_vitB16_imagenet}
\end{table}

\begin{table}[!ht]
\setlength{\tabcolsep}{3.0pt}
\centering
\scriptsize
\begin{tabular}{lcccccccc}
\toprule
\multicolumn{1}{c}{\multirow{2}{*}{Method}} & & \multicolumn{3}{c}{$\alpha=0.10$} &  & \multicolumn{3}{c}{$\alpha=0.05$}   \\ \cmidrule{3-5}\cmidrule{7-9}  
\multicolumn{1}{c}{}      & Top-1$\uparrow$       & Cov   & Size$\downarrow$ & CCV$\downarrow$ & & Cov.   & Size$\downarrow$ & CCV$\downarrow$  \\
\midrule 
\multicolumn{9}{l}{\textbf{SUN397}} \\ \hdashline
LAC \cite{lac}                                   & 76.0 & 0.900 & 1.69 & 9.12  &  & 0.951 & 2.70 & 5.26 \\
\our \hspace{2mm}w/ Conf-OT                      & \our77.4 & \our0.900 & \our1.59 & \our7.96  & \our & \our0.950 & \our2.44 & \our4.81 \\
APS \cite{aps}                                   & 76.0 & 0.899 & 3.22 & 6.53  &  & 0.950 & 5.60 & 4.36 \\
\our \hspace{2mm}w/ Conf-OT                      & \our77.4 & \our0.898 & \our2.77 & \our6.00  & \our & \our0.949 & \our4.62 & \our4.13 \\
RAPS \cite{raps}                                 & 76.0 & 0.898 & 3.03 & 6.58  &  & 0.949 & 4.78 & 4.37 \\
\bottomrule
\end{tabular}
\vspace{-2mm}
\caption{Results on fine-grained tasks with MetaCLIP ViT-H/14.}
\label{appendix_results_MetaCLIP_vith14_10tasks}
\end{table}

\begin{table}[!ht]
\setlength{\tabcolsep}{3.0pt}
\centering
\scriptsize
\begin{tabular}{lcccccccc}
\toprule
\multicolumn{1}{c}{\multirow{2}{*}{Method}} & & \multicolumn{3}{c}{$\alpha=0.10$} &  & \multicolumn{3}{c}{$\alpha=0.05$}   \\ \cmidrule{3-5}\cmidrule{7-9}  
\multicolumn{1}{c}{}      & Top-1$\uparrow$       & Cov   & Size$\downarrow$ & CCV$\downarrow$ & & Cov.   & Size$\downarrow$ & CCV$\downarrow$  \\
\midrule 
\multicolumn{9}{l}{\textbf{ImageNet}} \\ \hdashline
LAC \cite{lac}                                   & 80.7 & 0.901 & 1.42  & 8.84  &  & 0.950 & 2.33  & 5.38 \\
\our \hspace{2mm}w/ Conf-OT                      & \our80.6 & \our0.900 & \our1.42  & \our7.99  & \our & \our0.950 & \our2.36  & \our4.89 \\
APS \cite{aps}                                   & 80.7 & 0.899 & 5.60  & 6.64  &  & 0.949 & 12.37 & 4.58 \\
\our \hspace{2mm}w/ Conf-OT                      & \our80.6 & \our0.898 & \our4.24  & \our6.22  & \our & \our0.950 & \our8.87  & \our4.30 \\
RAPS \cite{raps}                                 & 80.7 & 0.899 & 3.90  & 6.68  &  & 0.949 & 6.15  & 4.66 \\
\bottomrule
\end{tabular}
\vspace{-2mm}
\caption{Results on ImageNet shifts with MetaCLIP ViT-H/14.}
\label{appendix_results_MetaCLIP_vith14_imagenet}
\end{table}

\end{document}